\documentclass[journal]{IEEEtran}

\usepackage{array}
\usepackage{cite}

\usepackage{algorithm}
\usepackage{algorithmic}
\usepackage{mdwmath}

\usepackage{graphicx}        
\usepackage{multicol}        
\usepackage{subfigure}
\usepackage{pifont}

\usepackage{bm}
\usepackage{comment}
\usepackage{amsmath}
\usepackage{enumerate}  
\usepackage{mathrsfs}
\usepackage{makeidx}         
\usepackage{graphicx}        
\usepackage{multicol}        
\usepackage{subfigure}
\usepackage{epsfig,amssymb,latexsym}
\usepackage{psfrag}
\usepackage{algorithmic}
\usepackage{algorithm}

\usepackage{fancyhdr}
\usepackage{multirow}

\usepackage{color}
\usepackage{indentfirst}

\renewcommand{\algorithmicrequire}{\textbf{Input:}}   
\renewcommand{\algorithmicensure}{\textbf{Output:}}   

\newcommand{\mr}{\mathrm}
\newcommand{\mc}{\mathcal}

\begin{document}

\title{Weakly-Supervised Cross-Domain Adaptation for Endoscopic Lesions Segmentation}

\author{
	Jiahua~Dong,
	Yang~Cong,~\IEEEmembership{Senior~Member,~IEEE,}
	Gan~Sun,
	Yunsheng~Yang,
	Xiaowei~Xu,
	and Zhengming Ding

	\thanks{This work is supported by Ministry of Science and Technology of the People´s Republic of China (2019YFB1310300), National Nature Science Foundation of China under Grant (61722311, U1613214, 61821005, 61533015) and National Postdoctoral Innovative Talents Support Program (BX20200353). \textit{(Corresponding author: Yang Cong.)}}
	
	\thanks{Jiahua Dong is with the State Key Laboratory of Robotics, Shenyang Institute of Automation, Chinese Academy of Sciences, Shenyang 110016, China, and also with the Institutes for Robotics and Intelligent Manufacturing, Chinese Academy of Sciences, Shenyang 110016, China, and also with the University of Chinese Academy of Sciences, Beijing 100049, China (e-mail: dongjiahua@sia.cn).}
	
	\thanks{Yang Cong and Gan Sun are with the State Key Laboratory of Robotics, Shenyang Institute of Automation, Chinese Academy of Sciences, Shenyang 110016, China, and also with the Institutes for Robotics and Intelligent Manufacturing, Chinese Academy of Sciences, Shenyang 110016, China (e-mail: congyang81@gmail.com, sungan1412@gmail.com).}
	
	\thanks{Yunsheng Yang is with the Chinese PLA General Hospital, Beijing 100000, China (email: sunny301ddc@126.com).}
	
	\thanks{Xiaowei Xu is with the Department of Information Science, University of Arkansas at Little Rock, Arkansas 72204, USA (e-mail: xwxu@ualr.edu).}
	
	\thanks{Zhengming Ding is with the Department of Computer, Information and Technology, Indiana University-Purdue University Indianapolis, Indianapolis, IN 46202 USA (email: zd2@iu.edu).}
	
}

\markboth{IEEE Transactions on Circuits and Systems for Video Technology,~Vol.~14, No.~8, August~2020}
{Shell \MakeLowercase{\textit{Dong et al.}}: Bare Demo of IEEEtran.cls for IEEE Journals}

\maketitle

\begin{abstract}
Weakly-supervised learning has attracted growing research attention on medical lesions segmentation due to significant saving in pixel-level annotation cost. However, 1) most existing methods require effective prior and constraints to explore the intrinsic lesions characterization, which only generates incorrect and rough prediction; 2) they neglect the underlying semantic dependencies among weakly-labeled target enteroscopy diseases and fully-annotated source gastroscope lesions, while forcefully utilizing untransferable dependencies leads to the negative performance. To tackle above issues, we propose a new weakly-supervised lesions transfer framework, which can not only explore transferable domain-invariant knowledge across different datasets, but also prevent the negative transfer of untransferable representations. Specifically, a Wasserstein quantified transferability framework is developed to highlight wide-range transferable contextual dependencies, while neglecting the irrelevant semantic characterizations. Moreover, a novel self-supervised pseudo label generator is designed to equally provide confident pseudo pixel labels for both hard-to-transfer and easy-to-transfer target samples. It inhibits the enormous deviation of false pseudo pixel labels under the self-supervision manner. Afterwards, dynamically-searched feature centroids are aligned to narrow category-wise distribution shift. Comprehensive theoretical analysis and experiments show the superiority of our model on the endoscopic dataset and several public datasets.
	
\end{abstract}

\begin{IEEEkeywords}
Weakly-supervised learning, endoscopic lesions segmentation, semantic knowledge transfer, domain adaptation.
\end{IEEEkeywords}

 \ifCLASSOPTIONpeerreview
 \begin{center} \bfseries EDICS Category: 3-BBND \end{center}
 \fi
%
\IEEEpeerreviewmaketitle

\section{Introduction}
\IEEEPARstart{W}{eakly-supervised} learning has been successfully extended into various computer vision tasks, especially for expensive tasks in terms of pixel-level annotations, \emph{e.g.,} semantic segmentation of medical images~\cite{exp:CDWS, Weak_super_1, Dong_2019_ICCV}. It has made great contributions to assisting the clinical experts by learning a semantic lesions segmentation model for pixel-level predictions with only accessing weakly-annotated (image-level) labels. To improve the accuracy and efficiency of medical lesions diagnosis, it has been successfully applied into a large amount of clinical diagnosis tasks until now, \emph{e.g.,} thoracic disease localization \cite{yan2018weakly}, automated glaucoma detection \cite{zhao2019weakly}, histopathology segmentation \cite{exp:CDWS}, etc.

\begin{figure}[t]
	\small
	\centering
	\includegraphics[trim = 0mm 0mm 0mm 0mm, clip, width=248pt, height  =125pt]{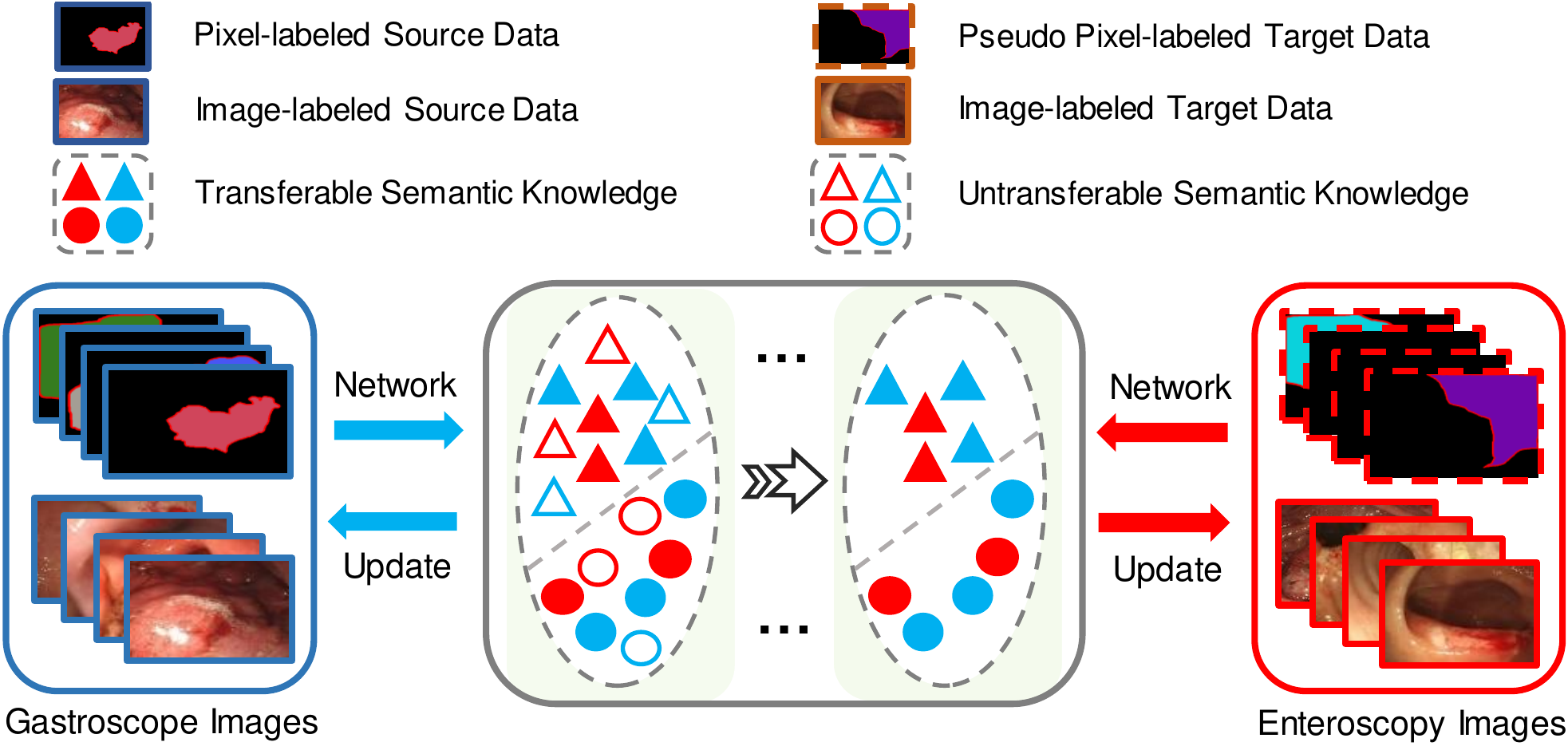}
	\caption{Illustration of our weakly-supervised semantic knowledge transfer framework. It selectively highlights transferable semantic characterizations from source lesions to assist the target diseases segmentation while neglecting these untransferable representations.}
	\label{figure:demonstration}
\end{figure}

However, when required to learn lesions locations by providing image category supervision, weakly-supervised semantic segmentation of medical lesions is still a thorny diagnosis task since: 1) prior distribution of lesions and clinical expertise \cite{exp:CDWS} are required to directly associate high-level semantic representation with low-level appearance, which cannot produce discriminative and correct lesions locations; 2) it neglects relevant semantic dependencies among target weakly-labeled diseases and source finely-labeled lesions. Obviously, semantic knowledge between two different domains (\emph{e.g.}, enteroscopy and gastroscope datasets) may be not all equally transferable, while forcefully adapting the untransferable dependencies is vulnerable to negative transfer of irrelevant knowledge. Furthermore, it is thorny to manually determine what kinds of semantic dependencies are transferable or not. For example, endoscopic lesions with various appearances and different backgrounds in the images could not contribute equally to semantic transfer in the feature space. Therefore, in this paper, how to establish a weakly-supervised semantic lesions transfer model with highly-selective capacity for transferable knowledge while ignoring irrelevant semantic knowledge among different domains is our main focus.

To tackle above challenges, as depicted in Fig.~\ref{figure:demonstration}, we propose a new weakly-supervised lesions transfer model, whose goal is to exploit transferable semantic knowledge between fully-labeled source gastroscope domain and weakly-labeled target enteroscopy domain, and brush irrelevant knowledge aside in a highly-selective manner.
To be specific, a Wasserstein adversarial learning based quantified transferability framework guides our model to focus on important semantic knowledge with high transfer scores, while ignoring these irrelevant dependencies. It incorporates two types of complementary parallel self-attention mechanisms to adaptively capture long-range contextual dependencies, and aligns the heterogeneous probability distributions among different datasets under the Wasserstein distance metric. In contrast to \cite{Dong_2019_ICCV}, a self-supervised pseudo pixel label generator under curriculum learning strategy could produce confident pseudo labels by taking into account class balance and super-pixel prior \cite{SLIC}. It pays equal attention to both hard-to-transfer and easy-to-transfer classes of target dataset while preventing the enormous deviation of false pseudo labels and the gradual dominance of easy-to-transfer classes along the training process. Afterwards, in order to narrow the category-wise feature distribution shifts, we align the dynamically-searched feature centroids across domains, which are gradually updated with the assistance of previously learned features under the supervision of pseudo labels. In addition, theoretical analysis about how our model explores domain-invariant transferable knowledge across different datasets is elaborated. Empirical comparisons on endoscopic dataset and some representative non-medical datasets illustrate that our model improves the transfer performance significantly when compared with other state-of-the-art approaches.

The key contributions of this paper are presented as follows: 	
\begin{itemize}
	\setlength{\itemsep}{0pt}		
	\setlength{\parsep}{0pt}	
	\setlength{\parskip}{0pt}
												
	\item We propose a new weakly-supervised lesions transfer framework for  endoscopic lesions segmentation. To the best of our knowledge, this is an earlier attempt to selectively explore transferable representations while neglecting irrelevant lesions characterizations in the biomedical image processing field. 
	
	\item We develop a Wasserstein quantified transferability framework based on two types complementary parallel self-attention mechanisms to selectively transfer important knowledge with high transferability while ignoring those untransferable semantic dependencies.
	
	\item A self-supervised pseudo label generator is designed to  equally generate confident pseudo pixel labels for both hard-to-transfer and easy-to-transfer classes of target dataset, which can not only inhibit gradual dominance of easy-to-transfer classes along the training process, but also further prevent the enormous deviation of false pseudo labels. 
	
	\item Comprehensive theoretical analysis guarantees that our model could effectively explore domain-invariant knowledge to minimize the domain discrepancy. 
	
\end{itemize}

This work is an extension version of our previous conference paper \cite{Dong_2019_ICCV} and the main improvements include: 1) A Wasserstein quantified transferability framework is developed to highlight important transferable dependencies while neglecting these untransferable semantic knowledge. It mitigates the heterogeneous probability distributions shift among different datasets under the Wasserstein distance metric. 2) Both spatial attention network and channel attention network are exploited to capture long-range context information for endoscopic lesions with various appearances. 3) Different from our conference version \cite{Dong_2019_ICCV}, a self-supervised pseudo label generator is developed to progressively produce more confident pseudo labels under a self-supervision way, which is determined by cross-entropy loss rather than probability \cite{Dong_2019_ICCV}. It further prevents the enormous deviation of false pseudo labels in the fine-tuning process. 4) Theoretical analysis guarantees that our model could effectively bridge the cross-domain distribution shift to learn transferable knowledge.

\section{Related Work} \label{sec:related work}
This section detailedly introduces some related researches about semantic segmentation of medical lesions and semantic knowledge transfer.

\subsection{Semantic Segmentation of Medical Lesions}
In the past few decades, computer aided diagnosis (CAD) \cite{cad_2} has been widely applied to assist clinician due to its significant improvement in the accuracy and efficiency of medical lesions segmentation. Traditional methods \cite{traditional:seg1, traditional:seg2} rely on clinical experience \cite{traditional:seg3} and domain experts \cite{traditional:seg4} to handcraft local image features to characterize medical lesions. After deep convolutional neural networks \cite{related:resnet, wang2019laplacian, wang2020recurrent} achieve great successes, many advanced methods \cite{medical_segment_springer, medical_seg_fcn} relying on powerful nets are introduced to attain better performance for semantic segmentation task of medical lesions. They require large-scale finely-annotated pixel labels to train the whole network architectures.
In order to efficiently save pixel annotation efforts, weakly-supervised semantic segmentation \cite{10.1109/ICCV.2015.209} focuses on exploring the dense pixel-wise labeling from image-level tags. Specifically, \cite{10.1109/ICCV.2015.209} proposes a novel loss function to optimize the output probability space with any set of linear constraints. Inspired by \cite{10.1109/ICCV.2015.209}, weakly-supervised learning for medical lesions segmentation task \cite{exp:CDWS, Weak_super_1, yan2018weakly} has attracted growing research interests.
Generally, they require effective prior \cite{yan2018weakly} and constraints \cite{exp:CDWS, zhao2019weakly} to explore discriminative lesions representation associated with image-level labels while producing inaccurate and coarse lesions localization. \cite{KERVADEC201988} introduces a differentiable penalty for the loss function to avoid expensive Lagrangian dual iterations and proposal generation. Basically, there is still a significant prediction gap between the training models with pixel-level supervision and image-level guidance.

\subsection{Semantic Knowledge Transfer}
After \cite{domain:class-Ganin-1} introduces generative adversarial network \cite{intro:gan-2} to minimize feature distribution gap across domains, various variants \cite{Kurmi_2019_CVPR, 8746823} are proposed for classification adaptation task. As pointed out by Zhang et al. \cite{exp:CL}, the domain adaptation methods for classification cannot be directly applied into the semantic segmentation task. To this end, \cite{domain:transfer} proposes to translate the source data to target data for domain adaptation in segmentation task. Several researches \cite{exp:Wild, exp:CCA, domain:class-preserve-2, Wu_2018_ECCV, Saito_2018_CVPR, exp:LSD, exp:CGAN, Xia_2020_CVPR} focus on utilizing generative adversarial architecture to transfer domain-invariant knowledge across domains in the feature space. Besides, Tsai \emph{et al.} \cite{exp:LtA} develop a multi-level contextual adaptation framework in the output space. Different from them, Zou et al. \cite{Zou_2018_ECCV, Zou_2019_ICCV} propose a self-training strategy instead of adversarial learning for semantic representation transfer. 
\cite{Wang2019WeaklySA} considers the bounding box as weak supervision to  guide cross-domain urban scenes adaptation.
\cite{Gong_2019_CVPR} translates source dataset into a continuous sequence of intermediate datasets with different style to smoothly bridge domain gap. \cite{Lee_2019_CVPR, Vu_2019_CVPR, Luo_2019_CVPR, CSCL_Dong_ECCV2020} propose some novel adaptation losses to narrow the distribution discrepancy or semantic inconsistency across domains. \cite{Dong_2020_CVPR} develops a quantified transferability perception mechanism to highlight transferable representations across gastroscope and enteroscopy domains, while neglecting untransferable knowledge. 
\cite{article_Deng} introduces a similarity guided constraint to map features from the same class across domains nearby. \cite{8946732} proposes an importance sampling mechanism to quantify sample contributions, which effectively prevents the distractions from the unexpected noisy samples. \cite{8370105} integrates the category-wise matching strategy into a transferable feature learning module and a robust task predictor.

\begin{figure*}[t]  
	\small
	\centering
	\includegraphics[trim = 0mm 0mm 0mm 0mm, clip, width =515pt, height =145pt]{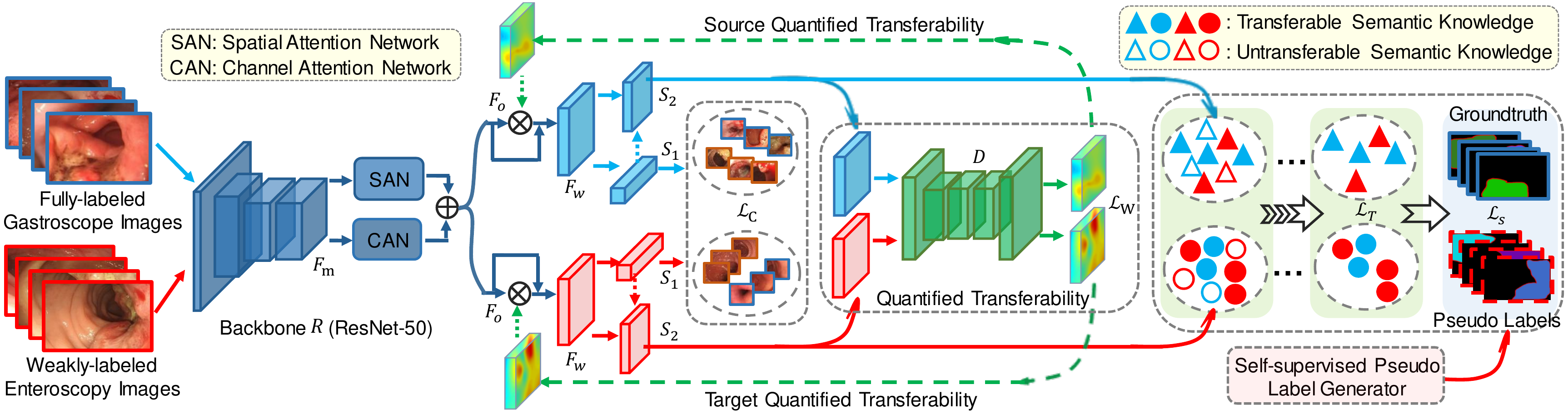}
	\caption{Overview of our proposed model that mainly contains three significant modules, \emph{i.e.}, \textit{a Wasserstein adversarial learning based quantified transferability} $\mc{L}_W$ for exploring transferable lesions dependencies while ignoring irrelevant knowledge, \textit{a self-supervised pseudo label generator} to produce confident pseudo labels for weakly-labeled target samples, and \textit{a semantic consistent loss} $\mc{L}_T$ to align category-wise feature centroids across domains.} 
	\label{figure:overview_algorithm}
\end{figure*}

\section{The Proposed Model} \label{sec:model_formulation}
In this section, we present brief model overview and detailed formulation of our weakly-supervised lesions transfer framework, followed by the theoretical analysis.

\subsection{Overview}				
As the overall framework demonstrated in Fig.~\ref{figure:overview_algorithm}, we respectively design two subnets denoted as $S_1$ and $S_2$ for classification and segmentation, where classification probability generated by $S_1$ is used to refine the pixel prediction of subnet $S_2$ via convolution operation. Denote the source gastroscope samples and target enteroscopy data as $X^s=\{(x_i^s,y_{i}^{sc},y_{i}^{ss})\}_{i=1}^{n_s} \in P_s$ and  $X^t=\{(x_j^t,y_{j}^{tc})\}_{j=1}^{n_t} \in P_t$, respectively, where $y_{i}^{sc}$ and $y_{i}^{ss}$ represent the image-level and pixel-level annotations of source image $x_i^s$, and $y_{j}^{tc}$ indicates the image tag of target image $x_j^t$. $n_s$ and $n_t$ represent the numbers of images for each domain. $P_s$ and $P_t$ represent the source and target marginal distributions. The image $x_i^s$ of source dataset $X^s$ is first passed into the whole architecture excluding discriminator $D$ to optimize network parameters. We then utilize the backbone ResNet-50 to extract semantic features for image $x_j^t$ of target dataset $X^t$. Discriminator $D$ takes feature maps of source dataset $X^s$ and target dataset $X^t$ as input to identify which one is from $X^s$ or $X^t$, which encourages both $X^s$ and $X^t$ to share closer feature distribution in the high-level semantic space.

Even though generative adversarial learning is employed to mitigate the domain gap between $X^s$ and $X^t$, it cannot ensure the category-wise feature distributions across domains are mapped nearby. To this end, we focus on aligning category-wise feature centroids to narrow the distribution shift. Unfortunately, feature centroids for target dataset $X^t$ are difficult to compute due to lack of valid pixel annotations. To tackle this challenge, we develop a novel self-supervised pseudo pixel labels generator that incorporates class balance and super-pixel prior. Relying on the generated pseudo pixel labels for target data $X^t$, we compute semantic centroids for each class by resorting to previously learned features. However, the semantic representations between gastroscope and enteroscopy domains are not all transferable, while forcefully exploiting these untransferable representations could lead to negative performance. Therefore, as depicted in Fig.~\ref{figure:overview_algorithm}, a Wasserstein adversarial network based quantified transferability framework is designed to explore transferable dependencies in both spatial and channel dimensions, while neglecting the irrelevant representation.

\subsection{Model Formulation} \label{sec:Model_Formulation}
To explore transferable knowledge for target lesions segmentation task in a highly-selective manner, our proposed model is formulated as follows:	
\begin{equation}
\begin{aligned}
\mathcal{L}_{} = &\mathcal{L}_C(X^s, X^t) + \mathcal{L}_S(X^s, X^t) + \\
&\eta\mathcal{L}_{W}(P_s, P_t) + \mu\mathcal{L}_T(X^s, X^t),
\end{aligned}		
\label{eqution:loss_all}		
\end{equation}
where $\eta\geq 0$ and $\mu\geq 0$ represent the balanced hyper-parameters. Each loss formulation in Eq.~\eqref{eqution:loss_all} can be defined as follows:	

\subsubsection{Classification Loss $\mathcal{L}_C(X^s, X^t)$}
The classification loss for both source gastroscope data and target enteroscopy data is expressed as $\mathcal{L}_C(X^s, X^t)$. The subnet $S_1$ is trained with loss $\mathcal{L}_C(X^s, X^t)$ to distinguish whether the input image has lesions regions or not. $\mathcal{L}_C(X^s, X^t)$ can be concretely written as:		
\begin{equation}
\begin{aligned}
\mathcal{L}_C(X^s, X^t) = &\mathbb{E}_{(x_i^s, y_{i}^{sc}) \in X^s}\big(- (y_{i}^{sc})^{\top}\mr{log}(S_1(x_i^s, \theta_{S_1}))\big)  \\
+&\mathbb{E}_{(x_j^t, y_{j}^{tc}) \in X^t}\big(-(y_{j}^{tc})^{\top}\mr{log}(S_1(x_j^t, \theta_{S_1}))\big),
\end{aligned}			
\label{equation:loss_classification} 								
\end{equation}
where $\theta_{S_1}$ indicates the network weights of the classifier $S_1$. $S_1(x_i^s, \theta_{S_1})$ and $S_1(x_j^t, \theta_{S_1})$ respectively denote the corresponding softmax outputs of classification network for source and target images.

\subsubsection{Segmentation Loss $\mathcal{L}_S(X^s, X^t)$} Given the predicted softmax outputs of the subnet $S_2$, we formulate $\mathcal{L}_S(X^s, X^t)$ as the segmentation loss for both source dataset $X^s$ with finely-annotated pixel-level labels $y_{i}^{ss}$, and target dataset $X^t$ with generated highly-confident pseudo pixel annotations. $\mathcal{L}_S(X^s, X^t)$ can then be expressed as:
\begin{equation}									
\begin{split}										
\mathcal{L}_S(&X^s, X^t)\! =\! \mathbb{E}_{(x_i^s, y_{i}^{ss}) \in X^s}\big(\!\!-\! \sum\limits_{a=1}^{|x_i^s|} (y_{ia}^{ss})^{\!\top} \log(S_2(x_i^s, \theta_{S_2})_a)\big)   \\
& + \mathbb{E}_{x_{j}^t \in X^t} \big(\!\! -\! \sum\limits_{b=1}^{|x_{j}^t|} v_j^b (\hat{y}_{jb}^{ts})^{\top} \log(S_2(x_j^t, \theta_{S_2})_b) \! - \! \lambda v_j^b \big),  \\   				
&s.t.,~\hat{y}_{jb}^{ts} \in \big\{ \mathbf{e}_k| \mathbf{e}_k \in \mathbb{R}^K, k = \arg\max\limits_{k} S_2(x_j^t, \theta_{S_2})_b \big\},  \\
&~~~~~~ v_j^b\in\{0, 1\},  ~\forall{b}=1,\ldots|x_{j}^t|,			
\end{split}	
\label{equation:loss_segmentation}				
\end{equation}	
where $\theta_{S_2}$ represents the network parameters of $S_2$. $S_2(x_i^s, \theta_{S_2})_a$ and $S_2(x_j^t, \theta_{S_2})_b$ respectively denote the classes probabilities predicted by pixel-wise classifier $S_2$ at pixel $a$ ($a=1,2,...,\left| x_i^s \right|$) and $b$ ($b=1,2,...,\left| x_j^t \right|$). 
$y_{ia}^{ss}$ and $\hat{y}_{jb}^{ts}$ are the one-hot groundtruth and confident pseudo label for the $a$-th and the $b$-th pixel positions in source image $x_i^s$ and target sample $x_j^t$, respectively. $K$ and $\mathbf{e}_k$ indicate the number of categories and one-hot vector. $v_j^b$ is the weight of the $b$-th pixel in target sample $x_j^t$. Notice that assigning $v_j^b$ as 0 can ignore its influence on  training process. Thus, we expect that $\lambda\geq 0$ can serve as a global trade-off controller to determine the selection scale of pseudo labels for target dataset, and prevent the trivial solution from assigning all generated pseudo labels as 0. Obviously, a larger $\lambda$ can promote the selection scale of confident pseudo pixel labels.

Although Eq.~\eqref{equation:loss_segmentation} could progressively mine pseudo pixel labels with high confidence score along the optimization process, the selection manner of pseudo labels in Eq.~\eqref{equation:loss_segmentation} still need to overcome three drawbacks: \textbf{(i)} Eq.~\eqref{equation:loss_segmentation} will have a preference for initially easier-to-learn classes while ignoring those hard-to-transfer categories in initial optimization procedure; \textbf{(ii)} The false pseudo labels could heavily cripple the performance along the training process; \textbf{(iii)} The selected pseudo pixel labels used for training are spatially discrete.

In order to cope with the issues \textbf{(i)} and \textbf{(ii)}, the second optimization term in Eq.~\eqref{equation:loss_segmentation} could be concretely rewritten as follows:
\begin{equation}
\begin{split}	
\min\limits_{v_b}&\mathbb{E}_{x_{j}^t \in X^t} \big(\!\! -\! \sum\limits_{b=1}^{|x_{j}^t|}\sum\limits_{k=1}^K p_j^b\big[v_j^b (\hat{y}_{jb}^{ts})_k^{\top} \log(S_2(x_j^t, \theta_{S_2})_b) \big] \! - \! \lambda_k v_j^b \big), \\
s.t&.,~\hat{y}_{jb}^{ts} \in \big\{ \mathbf{e}_k| \mathbf{e}_k \in \mathbb{R}^K, k = \arg\max\limits_{k} S_2(x_j^t, \theta_{S_2})_b \big\},  \\
p_j^b& = \max_{k} S_2(x_j^t, \theta_{S_2})_b, v_j^b\in\{0, 1\},  ~\forall{b}=1,\ldots |x_{j}^t|,
\end{split}
\label{equation:optimization_of_pseudo_label}
\end{equation}
where $\lambda_k$ ($k=1,2,...,K$) are class-balanced weights controlling selection scale of produced pseudo labels with respect to each class $k$. The cross-entropy loss of the $b$-th pixel is weighted by its maximum class probability $p_j^b$, which further averts the enormous deviation of false pseudo labels when retraining the whole network in a self-supervised manner. To prevent the dominance of large number of pixel categories, different from \cite{Dong_2019_ICCV}, cross-entropy loss of each pixel is employed to determinate the value of $\lambda_k$, as summarized in \textbf{Algorithm~1}. After computing cross-entropy loss $L_j$ for each pixel of all target images, we sort the losses $SL_k$ weighted by the predicted maximal probability of corresponding pixel for each class $k$. Then $\lambda_k$ can be determined by the self-supervised cross-entropy loss ranked at $n\mr{length}(SL_k)$. Besides, the value of $n$ starts from 25\% and increases by 5\% in each iterative optimization epoch, and we set the maximum value of $n$ as 55\%. The intuitive explanation about iteratively adding the value of $n$ is that the trained model will generate more confident pseudo labels as the training process. Moreover, the optimal solution of Eq.~\eqref{equation:optimization_of_pseudo_label} can be expressed as: 
\begin{equation}
\begin{split}
v_j^b =
\left\{
\begin{aligned}		
&1, \quad \mr{if} ~~(\hat{y}_{jb}^{ts})^{\top} \log(S_2(x_j^t, \theta_{S_2})_b) > - \frac{\lambda_k}{p_j^b},  \\
&0, \quad\quad\quad\quad \mr{otherwise}. \\
\end{aligned} 					
\right.											
\end{split}							
\label{equation:solution_of_pseudo_label}	
\end{equation}

To tackle the issue \textbf{(iii)}, the weight parameter $v_j = \{v_j^b\}_{b=1}^{\left|x_j^t\right|}$ of target image $x_j^t$ that are determined by Eq.~\eqref{equation:solution_of_pseudo_label} can be further modified by super-pixel prior \cite{SLIC}. It encourages spatial contextual continuity of selected pseudo labels for model retraining. Furthermore, the procedure about how to refine the weight parameter $v_j$ via super-pixel prior is elaborated in \textbf{Algorithm~2}, where $S_j^t$ represents super-pixel prior of target image $x_j^t$. When the weight parameter $v_j$ at the $(h, w)$-th pixel in target image $x_j^t$ equals to 0, its ultimate refinement can be determined by the information of its 8-neighborhoods under the voting strategy.

\renewcommand{\algorithmicrequire}{\textbf{Input:}}
\renewcommand{\algorithmicensure}{\textbf{Output:}}
\begin{algorithm}[t]			
	\caption{Determination Strategy of $\lambda_k$}
	\begin{algorithmic}[1]
		\REQUIRE The generation portion $n$ of pseudo labels, target enteroscopy sample $x_j^t \in X^t$, $K, S_2$;
		\ENSURE $\lambda_k$  \\
		\FOR {$j=1,\ldots,\left|X^t\right|$}
		\STATE  $Y_j = \mr{argmax}(S_2(x_j^t, \theta_{S_2}), \mr{axis}=3)$;
		\STATE  $M_j = \mr{max}(S_2(x_j^t, \theta_{S_2}),~\mr{axis}=3)$;
		\STATE  $L_j = \mr{pixel\_cross\_entropy\_loss}(Y_j, S_2(x_j^t, \theta_{S_2}))$;
		\STATE  $ML_j = M_j \circ L_j$, where $\circ$ denotes Hadamard product;
		\FOR {$k=1,\ldots,K$}
		\STATE  $ML_j^k = ML_j(Y_j == k)$;
		\STATE  $PL_k = [PL_k, \mr{vector}(ML_j^k)]$;
		\ENDFOR  \\
		\ENDFOR \\
		\FOR {$k=1,\ldots,K$}
		\STATE  $SL_k = \mr{ascending\_sorting}(PL_k)$;
		\STATE  $T_k = n\mr{length}(SL_k)$;
		\STATE  $\lambda_k = SL_k[T_k]$;
		\ENDFOR \\
		return $\lambda_k$;
	\end{algorithmic}
\end{algorithm}

\subsubsection{Wasserstein Adversarial Loss $\mathcal{L}_{W}(P_s, P_t)$} Due to the heterogeneous probability distribution of different datasets, Wasserstein-1 distance \cite{pmlr-v70-arjovsky17a} is employed to better measure the distribution discrepancy across $P_s$ and $P_t$ when comparing with other distance functions. The Wasserstein-1 discrepancy $\mathcal{L}_{W}(P_s, P_t)$ between $P_s$ and $P_t$ can be defined as follows:
\begin{equation}
\begin{split}
\!\!\mathcal{L}_{W}(P_s,\! P_t)\! =\!\inf \mathbb{E}_{(x_i^s, x_j^t)\in\Pi(P_s, P_t)}(|S_2(x_i^s, \theta_{S_2})\! -\! S_2(x_j^t, \theta_{S_2})|),
\end{split}
\label{equation:Wasserstein_Discrepancy}  	
\end{equation}
where $\Pi(P_s, P_t)$ denotes the joint distribution whose marginal distributions are $P_s$ and $P_t$. $S_2(x_i^s, \theta_{S_2})$ and $S_2(x_j^t, \theta_{S_2})$ respectively represent the extracted features for source and target samples. $\theta_{S_2}$ is the network weight of subnet $S_2$. Since the infimum of Eq.~\eqref{equation:Wasserstein_Discrepancy} is difficult to solve directly, $\mathcal{L}_{W}(P_s, P_t)$ in Eq.~\eqref{equation:Wasserstein_Discrepancy} is equally reformulated as below:
\begin{equation}
\begin{split}
\sup\limits_{||D\le 1||} \big|\mathbb{E}_{x_i^s\in P_s}\big(D(S_2(x_i^s, \theta_{S_2}))\big) - \mathbb{E}_{x_j^t\in P_t}\big(D(S_2(x_j^t, \theta_{S_2}))\big)\big|,
\end{split}
\label{equation:Wasserstein_Discrepancy_reformulation}  	
\end{equation}
where $D$ represents the domain critic function that takes the source and target features as input to identify which one is from $X^s$ or $X^t$. In this paper, we utilize $D$ as discriminator with the network parameter as $\theta_D$. Consequently, Eq.~\eqref{equation:Wasserstein_Discrepancy_reformulation} can be approximately rewritten as: 
\begin{equation}
\begin{split}
&\min\limits_{\theta_{S_2}}~\max\limits_{\theta_D} \mathcal{L}_W(P_s, P_t) =  \\ &\frac{1}{n_s}\sum\limits_{x_i^s\in X^s}D(S_2(x_i^s, \theta_{S_2}), \theta_D) - \frac{1}{n_t}\sum\limits_{x_j^t\in X^t}D(S_2(x_j^t, \theta_{S_2}), \theta_D) \\
& - \xi\big( (||\triangledown_{\theta_D}D(x_i^s, \theta_D)|| - 1)^2 - (||\triangledown_{\theta_D}D(x_j^t, \theta_D)|| - 1)^2\big),
\end{split}
\label{equation:Wasserstein_Discrepancy_loss_definition}  	
\end{equation}
where $\xi\geq0$ is the balanced weight.

However, the typical Wasserstein adversarial objective (\emph{i.e.}, Eq.~\eqref{equation:Wasserstein_Discrepancy_loss_definition}) cannot highlight transferable knowledge while ignoring the untransferable dependencies. Therefore, we develop a Wasserstein adversarial network based quantified transferability mechanism to selectively explore transferable semantic representation between $X^s$ and $X^t$. Specifically, as shown in Fig.~\ref{figure:overview_algorithm}, two types of complementary parallel self-attention networks (SAN and CAN) are integrated into generator (\emph{i.e.}, ResNet-50) to capture widely-separated contextual dependencies in both spatial and channel dimensions. Besides, discriminator $D$ is used to adaptively highlight long-range transferable representations and neglects irrelevant knowledge according to its output scores. The details about learning transferable dependencies for both generator and discriminator are shown as follows:

\renewcommand{\algorithmicrequire}{\textbf{Input:}}		
\renewcommand{\algorithmicensure}{\textbf{Output:}}		
\begin{algorithm}[t]	
	\caption{Selection Strategy of Weight Parameter $v_j$}
	\begin{algorithmic}[1]	
		\REQUIRE Height $I_h$ and width $I_w$ of target enteroscopy sample $x_j^t\in X^t$, $K$;
		\ENSURE Weight parameter $v_j$;   \\
		\STATE  Determine $\lambda_k$ via \textbf{Algorithm 1};
		\FOR {$j=1,\ldots,\left|X^t\right|~$}
		\STATE  Solve initial weight parameter $v_j$ of $x_j^t$ via Eq.~\eqref{equation:solution_of_pseudo_label};
		\STATE  Solve super-pixel prior $S_{j}^t$ of $x_j^t$ via \cite{SLIC};
		\FOR {$h=1,\ldots,I_h,~w=1,\ldots,I_w~$}
		\STATE  Set $C_{hw} = \emptyset$;
		\IF {$v_j$ equals to 0 at $(h,w)$-th pixel of $x_j^t$} 	
		\FOR {$k=1,\ldots,K~$} 	
		\STATE  $\!\!\!\!\!\!\!A=\{h-1, h, h+1\}, B=\{w-1, w, w+1\}$;	
		\STATE  $\!\!\!\!\!\!\!C_{hw}^k\!\! =\!\! \sum\limits_{x\in A}\!\sum\limits_{y\in B}\!\!\mathbf{1}_{((\hat{y}_{j}^{ts})_{xy} = k)\!\&((S_j^t)_{hw} = (S_j^t)_{xy})\!\&((v_j)_{xy}=1)}$;
		\STATE  $\!\!\!\!\!\!\!C_{hw} = [C_{hw}, ~C_{hw}^k]$;
		\ENDFOR
		\STATE  $N_k = \mr{argmax}(C_{hw}, ~\mr{axis}=0)$;
		\IF  {$C_{hw}[N_k]>4$}
		\STATE  $(v_j)_{hw} = 1$;
		\ENDIF
		\ENDIF
		\ENDFOR  \\ 	
		Return the ultimate weight parameter $v_j$ for $x_j^t$;
		\ENDFOR  \\		
	\end{algorithmic}					
\end{algorithm}

\begin{figure}[t]
	\centering
	\includegraphics[trim = 0mm 0mm 0mm 0mm, clip, width =245pt, height =205pt]{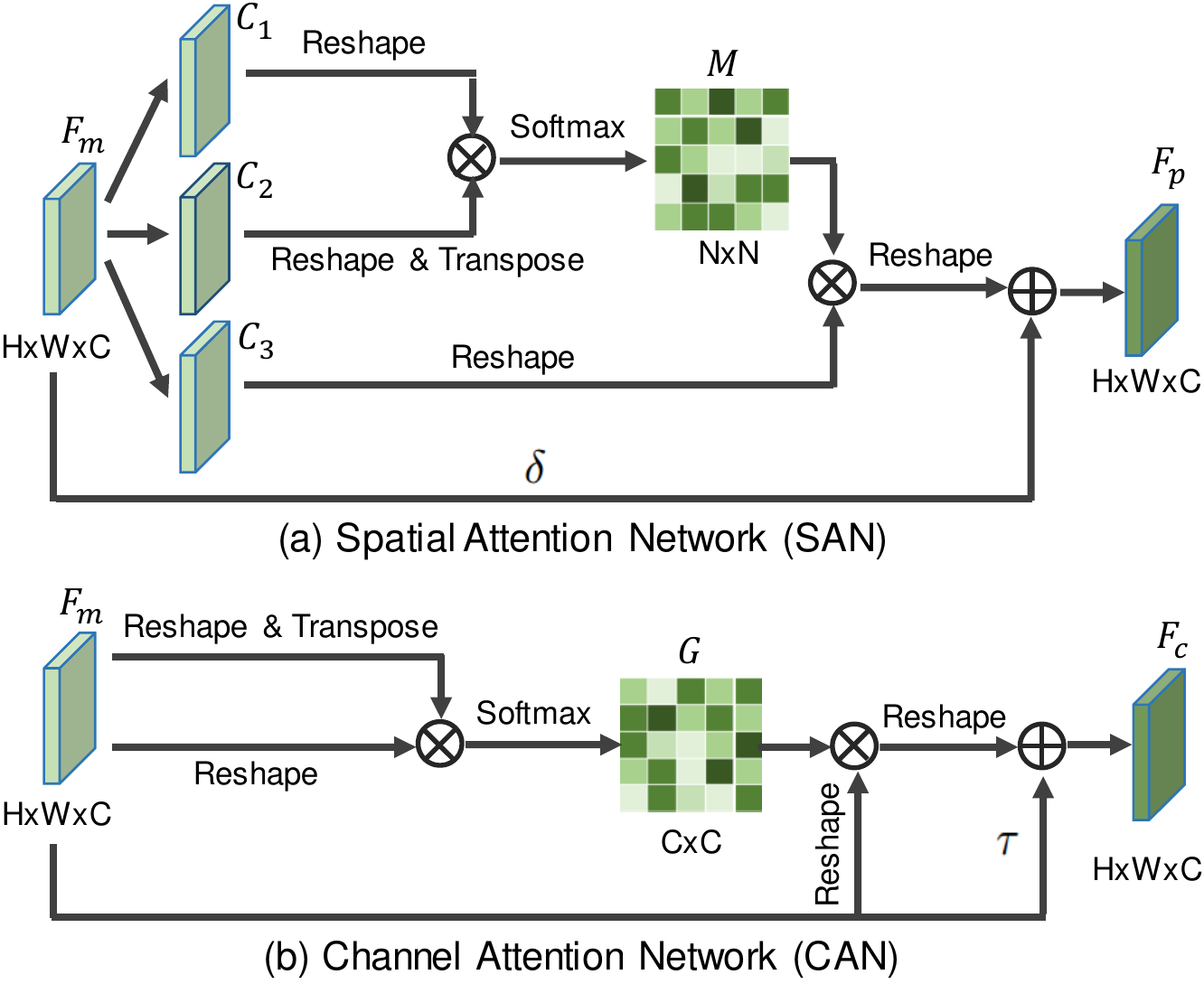}
	\caption{Detailed demonstration of (a) Spatial Attention Network (SAN) and (b) Channel Attention Network (CAN) in Fig.~\ref{figure:overview_algorithm}.}
	\label{fig:attention_module}
\end{figure}

\textbf{Generator Backbone $R$:} As illustrated in Fig.~\ref{figure:overview_algorithm}, we integrate both Spatial Attention Network (SAN) and Channel Attention Network (CAN) into the ResNet-50 \cite{related:resnet}, where the details about SAN and CAN are depicted in Fig.~\ref{fig:attention_module}.

For spatial attention network (SAN) demonstrated in Fig.~\ref{fig:attention_module} (a), given the intermediate feature $F_m\in\mathbb{R}^{H\times W\times C}$ extracted by ResNet-50 \cite{related:resnet}, we respectively feed it into two convolution layers to get two new features $C_1\in\mathbb{R}^{H\times W\times C}$ and $C_2\in\mathbb{R}^{H\times W\times C}$, where $H, W$ and $C$ indicate the height, width, channels of $F_m, C_1$ and $C_2$. After reshaping $C_1$ and $C_2$ into $\mathbb{R}^{N\times C}$ where $N = H\times W$ represents the total number of feature positions from hidden convolutional layer, we then apply a multiplication operation between $C_1$ and the transpose of $C_2$. Afterwards, softmax operator is utilized to calculate the position attention matrix $M=\{M_{ij}\}_{i,j=1}^N\in\mathbb{R}^{N\times N}$ and the $(i,j)$-th item of $M$ is formulated as:
$M_{ij} = \frac{\mr{exp}\big((C_1C_2^{\top})_{ij}\big)}{\sum_{i=1}^N \mr{exp}\big((C_1C_2^{\top})_{ij}\big)},
\label{equation:position_attention}$
where $M_{ij}$ measures the semantic representation correlation between different positions of corresponding feature map. The greater semantic dependency among them contributes to the larger value of $M_{ij}$.

Furthermore, feature $F_m$ is also forwarded into another convolutional block to produce a new feature $C_3\in\mathbb{R}^{H\times W\times C}$. We reshape it to $\mathbb{R}^{N\times C}$, and perform a multiplication operation among reshaped $C_3$ and the transpose of $M$ to output the attention feature map. Afterwards, we multiply the reshaped attention feature map ($\in\mathbb{R}^{H\times W\times C}$) by a scalar weight $\delta$, and conduct an element-wise sum operation with $F_m$ to output $F_p\in\mathbb{R}^{H\times W\times C}$:
\begin{equation}
(F_p)_i = \delta\sum_{j=1}^N M_{ij}(C_3)_j + (F_m)_i,
\label{equation:position_attention_feature}
\end{equation}
where $(F_p)_i, (F_m)_i$ and $(C_3)_j$ respectively represent the semantic representations at the $i$-th and $j$-th locations in the corresponding feature maps. Motivated by \cite{Zhang_attention_GAN}, we initialize $\delta$ as 0 and its value is learned adaptively.

Furthermore, for channel attention network (CAN) depicted in Fig.~\ref{fig:attention_module} (b), channel attention $G\in\mathbb{R}^{C\times C}$ can be directly calculated with the input feature map $F_m\in\mathbb{R}^{H\times W\times C}$. Concretely, after reshaping feature $F_m$ to $\mathbb{R}^{N\times C}$, we employ a multiplication operator between the transpose of $F_m$ and $F_m$, followed by softmax operation to calculate the channel attention $G=\{G_{ij}\}_{i,j=1}^C\in\mathbb{R}^{C\times C}$, where
$G_{ij} = \frac{\mr{exp}\big((F_m^{\top}F_m)_{ij}\big)}{\sum_{i=1}^C \mr{exp}\big((F_m^{\top}F_m)_{ij}\big)}$ denotes interdependent correlation between different channels of feature map $F_m$. Besides, we utilize the matrix multiplication between $F_m$ and the transpose of channel attention $G$, and their result is then reshaped to $\mathbb{R}^{H\times W \times C}$. Afterwards, we multiply their reshaped matrix by a scale weight $\tau$, and utilize an element-wise sum operation with the original input $F_m$ to obtain ultimate feature map $F_c\in\mathbb{R}^{H\times W \times C}$:
\begin{equation}
(F_c)_i = \tau\sum_{j=1}^C G_{ij}(F_m)_j + (F_m)_i,
\label{equation:channel_attention_feature}
\end{equation}
where $(F_c)_i$ and $(F_m)_j$ represent the semantic representations at the $i$-th and the $j$-th locations in the feature maps $F_c$ and $F_m$, respectively. Similar to $\delta$ in  Eq.~\eqref{equation:position_attention_feature}, $\tau$ is an adaptive parameter and initialized as 0.

\textbf{Discriminator $D$:} Intuitively, the discriminative ability of domain discriminator $D$ could evaluate whether the representations across domains are transferable or untransferable. Take an example for illustration, the input features from different domains which are already aligned will confuse $D$ for discriminating which one is from $X^s$ or $X^t$. That is to say, the output probabilities of discriminator $D$ for domain recognition help us to easily distinguish whether the input feature map could be transferred or not. Consequently, to focus on those transferable semantic characterizations, uncertainty evaluation strategy (\emph{i.e}, entropy criterion $H(p) = - \sum_m p_m \mr{log}(p_m)$) is used to identify the transferability of each semantic representation. Given the output probability $D(S_2(x_i^s, \theta_{S_2}), \theta_D)$ of discriminator $D$, the quantified transferability for the source data $x_i^s$ could concretely expressed as: 
\begin{equation}
W = 1 - H(D(S_2(x_i^s, \theta_{S_2}), \theta_D)).
\label{equation:transferability}
\end{equation}
Similarly, the quantified transferability mechanism is also suitable for the target sample $x_j^t$, as shown in Fig.~\ref{figure:overview_algorithm}. However, incorrect transferability quantification may degrade the transfer performance significantly, so we develop a residual transferability quantification mechanism to mitigate such negative effect, \emph{i.e.},
\begin{equation}
F_w = (1+W)\circ F_o,
\label{equation:transferable_attention}
\end{equation}
where $\circ$ denotes Hadamard product. $F_w$ is shared feature map with quantified transferability, which captures transferable semantic representations, as presented in Fig.~\ref{figure:overview_algorithm}.

\subsubsection{Semantic Consistence Loss $\mathcal{L}_T(X^s, X^t)$}
To mitigate the feature distribution shift of same class across different datasets $X^s$ and $X^t$, we present $\mathcal{L}_T(X^s, X^t)$ to achieve semantic knowledge transfer, which aligns the feature centroids with previously-learned representations:
\begin{equation}	
\mathcal{L}_T(X^s, X^t) = \sum\limits_{k=1}^{K} \left\| C_k^s - C_k^t\right\|_2^2+\alpha\left\| C_k^s - C_k^t\right\|_1,	
\label{equation:loss_SRT}
\end{equation}
where $C_k^s$ and $C_k^t$ respectively denote the feature centroids of the $k$-th class in source and target domains. $\alpha\geq 0$ aims to balance the sparse feature property. Similarly to the exponential reward strategy in reinforcement learning \cite{RL-1}, we develop the \textbf{Algorithm~3} to compute feature centroids of each class with the assistance of exponentially-weighted previously learned features in each iteration. Meanwhile, the produced pseudo pixel labels via \textbf{Algorithm 2} are employed to conduct feature centroids alignment process.

The intuitive explanation for resorting previously-learned experience rather than directly aligning those newly calculated centroids is straightforward: 1) Category information among each mini-batch is not usually sufficient, \emph{e.g.}, some categories may be missing due to the random selection strategy; 2) The deviation of false pseudo pixel labels will greatly increase when the mini-batch size is small.

\renewcommand{\algorithmicrequire}{\textbf{Input:}}
\renewcommand{\algorithmicensure}{\textbf{Output:}}
\begin{algorithm}[t]
	\caption{Optimize Semantic Consistence Loss $\mathcal{L}_T$}
	\begin{algorithmic}[1]
		\REQUIRE $\{C_{k}^{s}\}_{k=1}^K, \{C_k^t\}_{k=1}^K$, MAX-ITER;
		\ENSURE $\mathcal{L}_{T}(X^s, X^t)$;  \\	
		\FOR {$t=1, 2, \ldots,$ MAX-ITER}
		\STATE  $\mathcal{L}_{T}(X^s, X^t) = 0$;
		\STATE  $\big((x_i^s, y_{i}^{ss}), (x_j^t)\big) = \mr{randomly\_sampling}(X^s, X^t)$;
		\STATE  $v_j = \mr{pseudo\_labeling}(x_j^t)$ via \textbf{Algorithm 2};
		\STATE  Extract pixel features $F_i^s$ and $F_j^t$ via $S_2$;	
		\FOR {$k=1,\ldots,K$}	
		\STATE  $C_{k}^{sn} = \frac{1}{\left|x_i^s \right|} \sum\limits_{a=1}^{\left|x_i^s \right|} (F_i^s)_{a} \textbf{1}_{(y_{i}^{ss})_{a}=k}$;
		\STATE  $C_{k}^{tn} = \frac{1}{\left|x_j^t \right|} \sum\limits_{b=1}^{\left|x_j^t \right|} (F_j^t)_{b} \textbf{1}_{(\hat{y}_{j}^{ts})_{b}=k}$;
		\STATE  $C_k^s\! =\! \sum_{x=1}^{n} C_{k}^{sx} \cdot \gamma ^{t-x}$; (Exponentially-weighted)
		\STATE  $C_k^t\! =\! \sum_{x=1}^{n} C_k^{tx} \cdot \gamma ^{t-x}$; (Exponentially-weighted)
		\ENDFOR   \\
		\STATE  Return $\mathcal{L}_{T}(X^s, X^t)$;
		\ENDFOR   \\
	\end{algorithmic}	
\end{algorithm}

\subsection{Details of Network Architecture}
\subsubsection{Backbone $R$, Subnet $S_1$ and $S_2$}
Following the framework presented in \cite{Dong_2019_ICCV}, we employ DeepLab-v3 \cite{net:deeplabv3} with ResNet-50 network \cite{related:resnet} as our baseline architecture. To output shared higher dimensional feature map for $S_1$ and $S_2$, we set the strides of the last two convolution layers of ResNet-50 as 1, and remove the last classification head. Besides, in the last convolution layer of ResNet-50, we adopt 3 dilated convolution kernel with the stride as \{1, 2, 4\} to expand receptive field. The shared semantic feature produced by backbone $R$ is then respectively forwarded into two parallel attention networks (\emph{i.e.}, SAN and CAN), which explores wide-range contextual dependencies via an element-wise feature fusion, as depicted in Fig.~\ref{figure:overview_algorithm}. Afterwards, we forward the fused feature into subnet $S_1$ for classification task, and highlight transferable dependencies via the transferability quantified by discriminator $D$ before passing it into subnet $S_2$ for pixel-level prediction. $S_2$ is composed of an Atrous Spatial Pyramid Pooling (ASPP) module \cite{intro:deeplab} and a pixel-level predictor.

\subsubsection{Discriminator $D$}
Motivated by \cite{DBLP:journals/corr/RadfordMC15}, for the main components of discriminator $D$, five fully convolution networks with kernel size as 3 are used for retaining global information when compared with fully connected network. Specifically, the channel numbers of 5 convolutional filters are \{16, 32, 64, 64, 1\}, respectively. We set the stride of first three convolutional filters as 2 and the stride of other layers as 1. The Leaky RELU function with the parameter as 0.2 is applied to activate each convolution block excluding the last one that is activated by sigmoid function.

\subsection{Training and Testing} \label{sec:Training_and_Testing}
\subsubsection{Training Phase} 
The overall objective of training our model is formulated as:
\begin{equation}
\begin{split}
\min\limits_{\theta_R, \theta_{S_1}, \theta_{S_2}} \max\limits_{\theta_D}~~\mathcal{L}_{obj} = \mathcal{L}_C +\mathcal{L}_S
 + \eta\mathcal{L}_{W} + \mu\mathcal{L}_{T}.
\end{split}
\label{equation:training_objective_overall}
\end{equation}

In each iteration step, we use ResNet-50 to extract feature maps of source (\emph{e.g.}, gastroscope) and target (\emph{e.g.}, enteroscopy) images, and then forward them into discriminator $D$ for optimizing $\mathcal{L}_W(P_s, P_t)$. To optimize classification loss $\mathcal{L}_C(X^s, X^t)$ and segmentation loss $\mathcal{L}_S(X^s, X^t)$, source sample $x_i^s$ with the transferability quantified by $D$ is passed into the subnets $S_1$ and $S_2$. Afterwards, we obtain the target softmax output $S_2(x_j^t, \theta_{S_2})$ for image $x_j^t$ only with the image-level annotation $y_{j}^{tc}$, and corresponding ultimate weight parameter $v_j$ is produced via \textbf{Algorithm~2}. Additionally, with assistance of previously learned features, the category feature centroids $C_k^s$ and $C_k^t$ are solved via \textbf{Algorithm~3} to optimize the loss $\mathcal{L}_{T}(X^s, X^t)$. As for detailed implementation, a Titan V GPU with 12 GB memory is utilized for training and the batch size is set as 4. We employ Adam as the optimizer with initial learning rate as $1.0\times10^{-4}$ to train the whole networks architecture. The rate and step size of Adam optimizer are respectively set as 0.7 and 950.

\subsubsection{Testing Phase} 
In the testing stage, we directly forward the target enteroscopy sample $x_j^t$ into the backbone network ResNet-50 followed by the subnets $S_1$ and $S_2$ for performance prediction.

\subsection{Theoretical Insights} \label{sec:theoretical_insights}
This subsection introduces comprehensive theoretical analysis about how our model could efficiently narrow the domain discrepancy between source distribution $P_s$ and target distribution $P_t$. According to \cite{BenDavid2010}, given the hypothesis class $\mathcal{H}$, the target expected error $\epsilon_T(h)$ of any classifier $h\in\mathcal{H}$ predicting on target domain can be theoretically bounded by the following formulation:
\begin{equation}
\begin{split}
\forall h\in\mathcal{H},~ \epsilon_T(h) \leq \epsilon_S(h) + \frac{1}{2}d_{\mathcal{H}\triangle\mathcal{H}(P_s, P_t)} + \Lambda,
\end{split}
\label{equation:theory_analysis}  	
\end{equation}
where $\epsilon_S(h)$ represents the source expected error of $h\in\mathcal{H}$ predicting on source domain, which could be easily minimized to 0 under the fully-supervised learning. $\Lambda$ denotes the independent constant. Thus, our model focuses on how to minimize the domain discrepancy $d_{\mathcal{H}\triangle\mathcal{H}(P_s, P_t)}$ between source images $x_i^s\in X^s$ and target samples $x_j^t\in X^t$, which can be concretely defined as:
\begin{equation}
\begin{split}
&d_{\mathcal{H}\triangle\mathcal{H}(P_s, P_t)} = 2\!\!\!\sup\limits_{h, h' \in\mathcal{H}} \!\!\!\! |{P_s}_{(h(x_i^s)\ne h'(x_i^s))} - {P_t}_{(h(x_j^t)\ne h'(x_j^t))}| \\
& = 2\!\!\! \sup\limits_{h, h' \in\mathcal{H}} \!\!\!\! |\mathbb{E}_{x_i^s\in X^s}(\mathbf{1}_{(h(x_i^s)\ne h'(x_i^s))}) - \mathbb{E}_{x_j^t\in X^t}(\mathbf{1}_{(h(x_j^t)\ne h'(x_j^t))})| \\
& = 2\!\!\! \max\limits_{h, h' \in\mathcal{H}} \!\!\! |\mathbb{E}_{x_i^s\in X^s}(\mathbf{1}_{(h(x_i^s)\ne h'(x_i^s))}) - \mathbb{E}_{x_j^t\in X^t}(\mathbf{1}_{(h(x_j^t)\ne h'(x_j^t))})|.
\end{split}
\label{equation:HtriangleH}  	
\end{equation}
The classifiers $h, h'\in\mathcal{H}~(h\ne h')$ in Eq.~\eqref{equation:HtriangleH} can be decomposed as the subnets $R \circ S_1$ and $R \circ S_2$. They perform on the high-level semantic features extracted from backbone network $R$ for classification and segmentation, respectively. With the fixed parameters of $R$, $S_1$ and $S_2$, Eq.~\eqref{equation:HtriangleH} can then be rewritten as Eq.~\eqref{equation:HtriangleH_maxCompose} by employing discriminator $D$ to distinguish different distributions, \emph{i.e.},
\begin{equation}
\begin{split}
\!\!\!\!d_{\mathcal{H}\triangle\mathcal{H}(P_s, P_t)} = 2&\max\limits_{\theta_D}~  \big\{|\mathbb{E}_{x_i^s\in X^s}(\mathbf{1}_{(R\circ S_1(x_i^s)\ne R\circ S_2(x_i^s))}) \\
&~~~ - \mathbb{E}_{x_j^t\in X^t}(\mathbf{1}_{(R\circ S_1(x_j^t)\ne R\circ S_2(x_j^t))})|\big\}.
\end{split}
\label{equation:HtriangleH_maxCompose}
\end{equation}
Eq.~\eqref{equation:HtriangleH_maxCompose} is in accordance with our  formulation in Eq.~\eqref{equation:Wasserstein_Discrepancy_loss_definition}, \emph{i.e.},
\begin{equation}
\begin{split}
&\max\limits_{\theta_D} \mathcal{L}_W(P_s, P_t) =  \\ &\frac{1}{n_s}\sum\limits_{x_i^s\in X^s}D(S_2(x_i^s, \theta_{S_2}), \theta_D) - \frac{1}{n_t}\sum\limits_{x_j^t\in X^t}D(S_2(x_j^t, \theta_{S_2}), \theta_D) \\
& - \xi\big( (||\triangledown_{\theta_D}D(x_i^s, \theta_D)|| - 1)^2 - (||\triangledown_{\theta_D}D(x_j^t, \theta_D)|| - 1)^2 \big).
\end{split}
\label{equation:HtriangleH_maxD}
\end{equation}

\begin{table*}[t]
	\centering
	\setlength{\tabcolsep}{1.17mm}
	\caption{Comparison experiments between our model and several state-of-the-art approaches on medical endoscopic dataset.}
	\scalebox{0.884}{
		\begin{tabular}{|c|cccccccccccc|c|}
			\hline
			Metrics & BL \cite{net:deeplabv3} & CDWS \cite{exp:CDWS} & NMD \cite{exp:CCA} & Wild \cite{exp:Wild} & DFN \cite{exp:DFN} & LtA \cite{exp:LtA} & CGAN \cite{exp:CGAN} & CLAN \cite{Luo_2019_CVPR} & ADV \cite{Vu_2019_CVPR} & CRST-LR \cite{Zou_2019_ICCV} & CRST-MR \cite{Zou_2019_ICCV} & SWE \cite{Dong_2019_ICCV} & ~~Ours~~ \\
			\hline
			\hline
			$\mr{IoU}_n$($\%$) & 75.13 &25.11 & 81.10 & 81.58& 81.33 & 81.73& 80.32 & 82.31 & 83.46 & 83.98 & 84.21 & 84.76 & \textbf{85.21} \\	
			
			$\mr{IoU}_d$($\%$) & 33.24 & 15.51 & 36.85 & 38.59 & 37.50 & 41.10 & 41.33 & 42.04 & 42.63& 42.85 & 42.94 & 43.16 & \textbf{43.65} \\
			
			mIoU($\%$) & 54.19 &20.31 &58.97 & 60.09 &59.41&61.42 & 60.82 & 62.18 & 63.05 & 63.42 & 63.57 & 63.96 & \textbf{64.43}  \\
			
			\hline					
		\end{tabular}
	}				
	\label{table:exp_medical_dataset}
\end{table*}

Considering the goal of minimizing the domain discrepancy $d_{\mathcal{H}\triangle\mathcal{H}(P_s, P_t)}$, we aim to mitigate the domain-wise distribution shift by optimizing $R, S_1$ and $S_2$ with the losses $\mathcal{L}_C(X^s, X^t), \mathcal{L}_S(X^s, X^t)$ and $\mathcal{L}_{T}(X^s, X^t)$, \emph{i.e.},
\begin{equation}
\begin{split}
\min\limits_{\theta_R, \theta_{S_1}, \theta_{S_2}} \max\limits_{\theta_D}~~&  \mathcal{L}_C(X^s, X^t) +\mathcal{L}_S(X^s, X^t) \\
& + \eta\mathcal{L}_{W}(P_s, P_t) + \mu\mathcal{L}_{T}(X^s, X^t).
\end{split}
\label{equation:theory_analysis_overall}
\end{equation}

Notice that Eq.~\eqref{equation:theory_analysis_overall} well matches our formulation in Section \ref{sec:Model_Formulation} and the training objective Eq.~\eqref{equation:training_objective_overall} in Section \ref{sec:Training_and_Testing}. Therefore, by iteratively updating the parameters of $R, S_1, S_2$ and $D$, our model could efficiently narrow the domain discrepancy $d_{\mathcal{H}\triangle\mathcal{H}(P_s, P_t)}$ and achieve the tighter upper bound for $\epsilon_T(h)$.

\section{Experiments} \label{sec:experiments}
This section first introduces several representative experimental datasets, and then reports comparison experiments with some state-of-the-art methods to justify the superiority of our model. 

\begin{table}[t]
	\centering
	\setlength{\tabcolsep}{1.8mm}
	\caption{Notations for different modules in our model and evaluation.}
	\scalebox{0.884}{
		\begin{tabular}{|c|c|}
			\hline
			Notations & Interpretation \\
			\hline
			BL & Baseline architecture DeepLab-v3 \cite{net:deeplabv3} without knowledge transfer \\
			PL & Pseudo labels generator \\
			AL & Wasserstein adversarial learning framework \\
			SRT & Semantic consistence loss $\mathcal{L}_T(X^s, X^t)$ in Eq.~\eqref{equation:loss_SRT}\\
			QT & Quantified transferability in Eq.~\eqref{equation:transferability} \\
			Ours-woSP & Training our model without the super-pixel prior\\
			\hline
			$\mr{mIoU}$ & mean intersection over union (IoU) \\
			$\mr{IoU}_d$ & intersection over union (IoU) of disease \\
			$\mr{IoU}_n$ & intersection over union (IoU) of normal \\
			\hline				
		\end{tabular}
	}				
	\label{table:exp_notations}
\end{table}

\subsection{Dataset and Notations}	
\textbf{Medical Endoscopic Dataset} \cite{Dong_2019_ICCV} is collected from more than 1100 volunteers, and contains 3659 images with various diseases, including cancer, polyp, gastritis, ulcer and bleeding. To be specific, it is composed of 2969 gasteroscope samples with pixel-level annotations and 690 enteroscopy samples only with image-level labels. All gastroscope data and 300 enteroscopy images are respectively regarded as source and target domains for model training. Besides, the rest of enteroscopy samples are employed for evaluation.

\textbf{Cityscapes} \cite{data:city} with 34 different scene categories is collected from fifty European cities. For training, validation and testing, there are three corresponding subsets with 2993, 1531, 503 samples, respectively.

\textbf{GTA} \cite{data:GTA} is composed of 24996 samples that are gathered from the computer game called Grand Theft Auto V. The annotation classes for Cityscapes \cite{data:city} and \textbf{GTA} \cite{data:GTA} are compatible with each other.

\textbf{SYNTHIA} \cite{data:synthia} is a large-scale synthetic dataset from virtual urban scenes, whose subset named SYNTHIA-RANDCITYSCAPES with total 9400 samples and 12 annotated object categories is employed for performance comparison.

\textbf{Notations:} For simplification, some notations used in this paper are summarized in Table~\ref{table:exp_notations}. To be specific, different components in our model, \emph{i.e.}, baseline architecture DeepLab-v3 \cite{net:deeplabv3}, pseudo labels generator, Wasserstein adversarial learning, semantic representation transfer, quantified transferability and training without super-pixel prior are abbreviated as BL, PL, AL, SRT, QT and Ours-woSP, respectively. 
Furthermore, we consider intersection over union (IoU) for basic metric evaluation, where mIoU, $\mr{IoU}_n$ and $\mr{IoU}_d$ respectively denote the mean IoU , IoU of normal and IoU of disease.

\begin{figure}[t]
	\centering
	\includegraphics[trim = 0mm 0mm 0mm 0mm, clip, width=255pt, height =160pt]{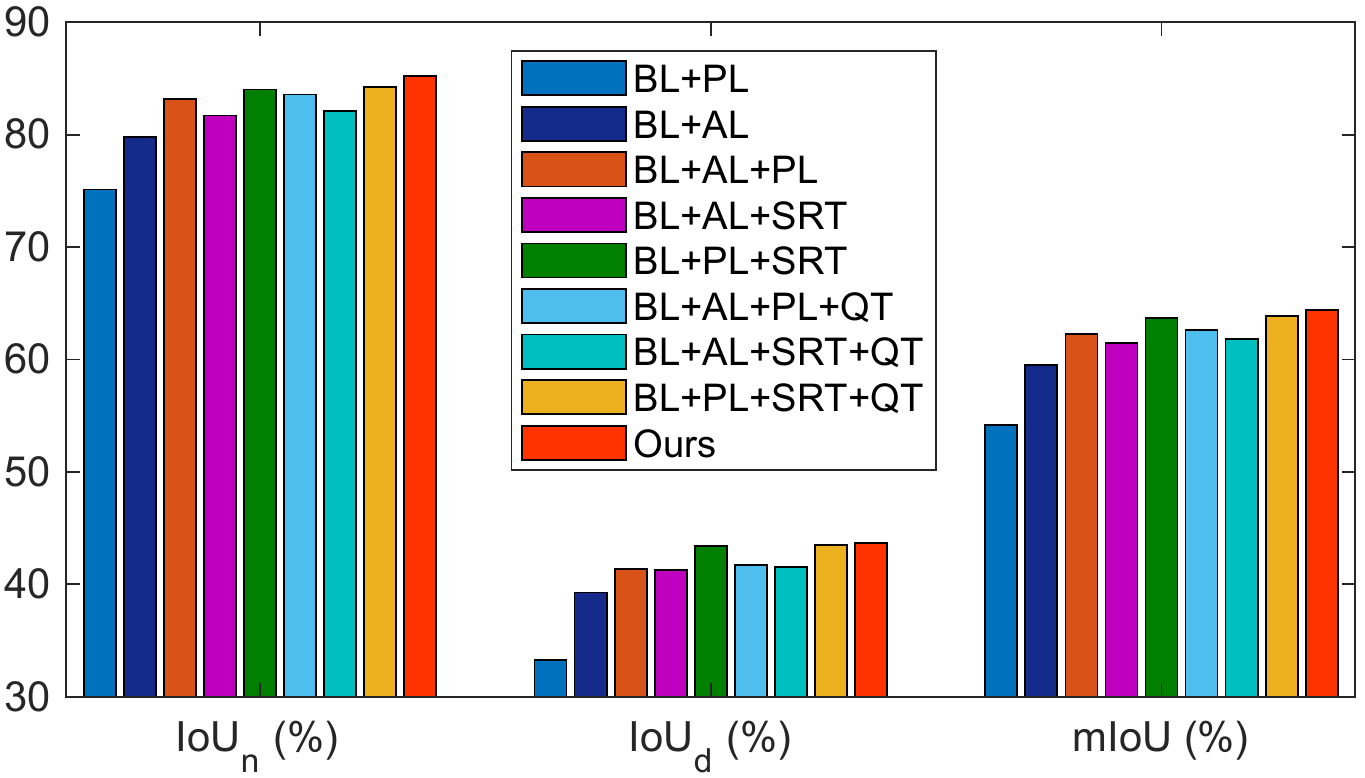}
	\caption{Ablation studies on medical endoscopic dataset in terms of $\mr{IoU}_n, \mr{IoU}_d$ and $\mr{mIoU}$ metrics.} 	
	\label{figure:ablation_exp_medical}	
\end{figure}

\subsection{Experiments on Medical Endoscopic Dataset}
Several state-of-the-art approaches are employed for experiment comparison, which are presented as follows:
\begin{itemize} 	
	\setlength{\itemsep}{0pt}
	\setlength{\parsep}{0pt}
	\setlength{\parskip}{0pt}
	
	\item \textbf{Baseline} (\textbf{BL}) employs DeepLab-v3 \cite{net:deeplabv3} as backbone for segmentation without knowledge transfer.
	
	\item \textbf{CDWS} \cite{exp:CDWS} explores weakly-supervised multi-scale model for pixel predictions by applying area constraint.
	
	\item \textbf{NMD} \cite{exp:CCA} integrates static object prior and soft pseudo labels into multiple class-wise adaptation.
	
	\item \textbf{Wild} \cite{exp:Wild} exploits intermediate convolutional blocks by utilizing a adversarial loss based on prior constraint.
	
	\item \textbf{DFN} \cite{exp:DFN} focuses on exploring discriminative semantic representations under the supervision manner. 
	
	\item \textbf{LtA} \cite{exp:LtA} develops a multi-level transfer model in the output space. 
	
	\item \textbf{CGAN} \cite{exp:CGAN} applies conditional GAN into feature space adaptation by incorporating conditional generator.
	
	\item \textbf{CLAN} \cite{Luo_2019_CVPR} exploits category-level alignment network.
	
	\item \textbf{ADV} \cite{Vu_2019_CVPR} proposes entropy minimization objectives  based on adversarial loss to narrow the domain gap.
	
	\item \textbf{CRST} \cite{Zou_2019_ICCV} focuses on relaxing the feasible space of pseudo-labels from one-hot vectors to continuous latent variables. We employ the label entropy regularizer (\emph{i.e.}, CRST-LR) and model regularizer $L_2$ (\emph{i.e.}, CRST-MR) for comparison.
	
	\item \textbf{SWE} \cite{Dong_2019_ICCV} is our conference version which develops a confident pseudo label generator to pick highly-confident samples into training set.  	
\end{itemize}

For performance comparison, we utilize DeepLab-v3 \cite{net:deeplabv3} with ResNet-50 \cite{related:resnet} as baseline network, and add an auxiliary image classifier to assist pixel predictions in this experiment. Table~\ref{table:exp_medical_dataset} reports the experimental results of our model against several competing approaches. Some observations from Table~\ref{table:exp_medical_dataset} are presented as follows: 
1) When compared with our conference version \cite{Dong_2019_ICCV}, the quantified transferability of our model plays an essential role in highlighting wide-range transferable dependencies for both normal and disease classes, even though the medical lesions across different datasets have different appearances and backgrounds. 2) Compared with \cite{Dong_2019_ICCV}, pseudo label generator can further prevent the enormous deviation of false pseudo labels in the fine-tuning phase when trained under the self-supervision manner. 
3) Our model outperforms the state-of-the-arts \cite{exp:LtA, exp:CGAN, Luo_2019_CVPR, Vu_2019_CVPR} by a large margin around 1.38$\sim$3.61\% in terms of mIoU since the self-supervised pseudo label generator could mine confident pseudo labels for target samples progressively. 4) Models with semantic transfer \cite{exp:CCA, exp:Wild, exp:DFN, exp:LtA, exp:CGAN, Luo_2019_CVPR, Vu_2019_CVPR} perform better than those without semantic transfer \cite{net:deeplabv3, exp:CDWS}.

\begin{figure}[t]
	\centering
	\includegraphics[trim = 0mm 0mm 0mm 0mm, clip, width =255pt, height =160pt]{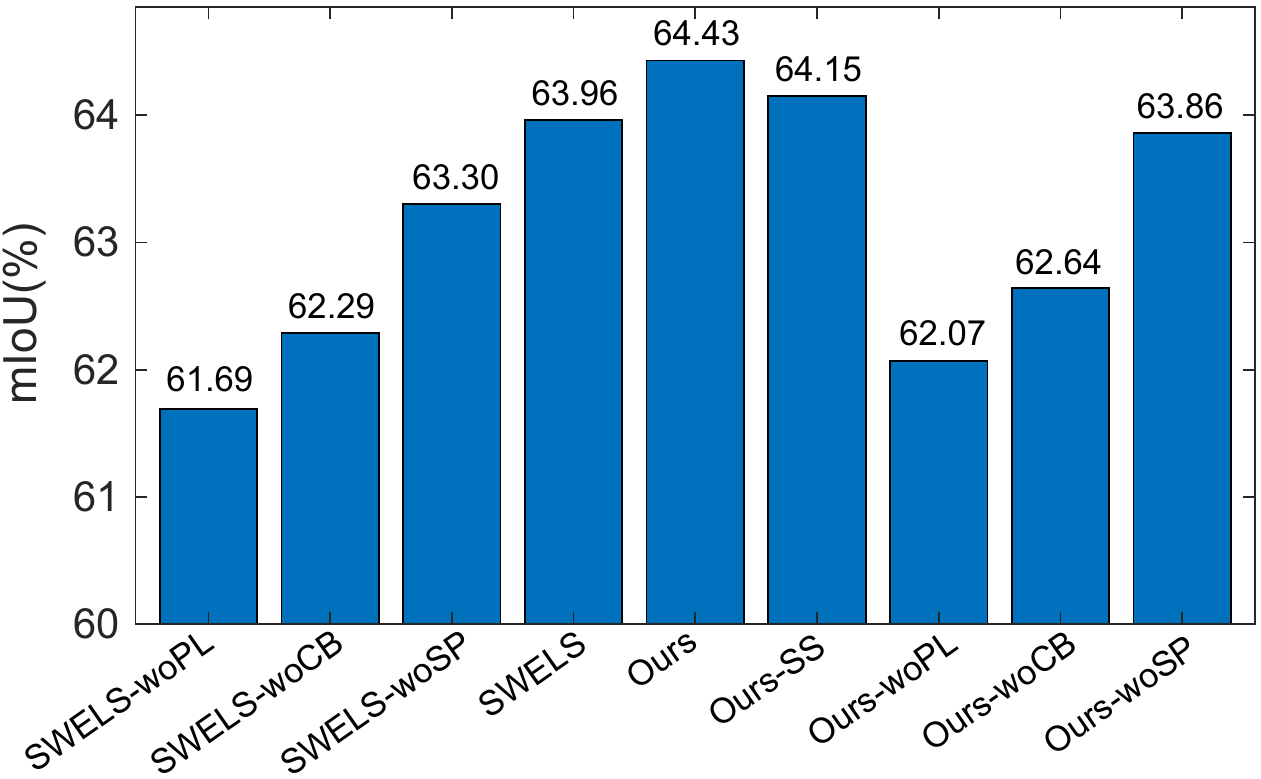}
	\caption{Discussion of different pseudo label designs on medical dataset.}
	\label{fig:ablation_pseudo_label}
\end{figure}

\begin{figure}[t]
	\centering
	\includegraphics[trim = 0mm 0mm 0mm 0mm, clip, width =252pt, height =95pt]{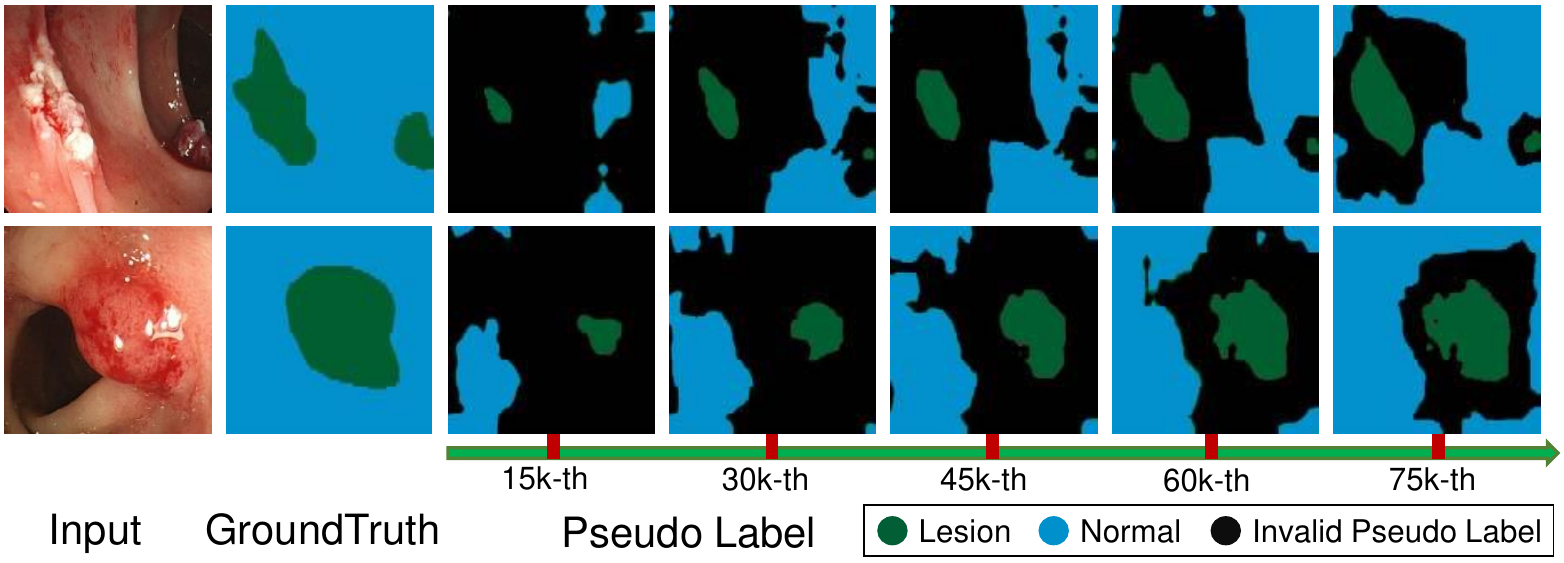}
	\caption{Examples of intuitively illustrating the propagation process of pseudo labels, where the inputs are enteroscopy samples.}
	\label{fig:pseudo_label_propagation_medical}
\end{figure}

\begin{figure}[t]
	\begin{minipage}[t]{0.495\linewidth}
		\centering
		\includegraphics[trim = 0mm 0mm 0mm 0mm, clip, height=3.8cm,width=4.4cm]{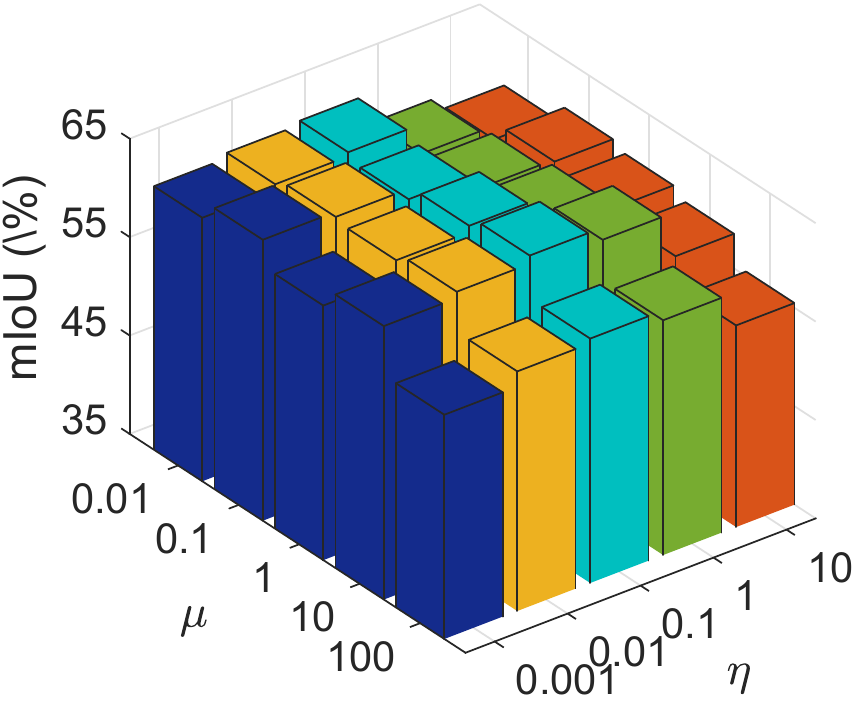}
		\\ {(a) when $\gamma=0.7, \alpha=1$}
		\label{fig:paras_a}
	\end{minipage}
	\begin{minipage}[t]{0.495\linewidth}
		\centering
		\includegraphics[trim = 0mm 0mm 0mm 0mm, clip, height=3.8cm,width=4.4cm]{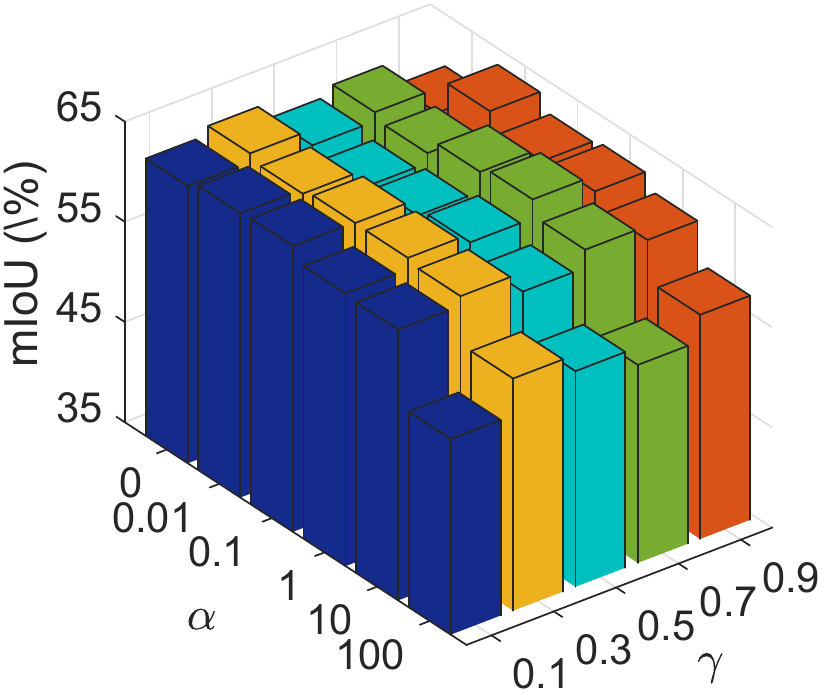}
		\\ {(b) when $\mu=10, \eta=0.3$}
		\label{fig:paras_b}
	\end{minipage}
	\caption{The parameter investigations with respect to \{$\mu, \eta$\} (a) and \{$\alpha, \gamma$\} (b) on medical endoscopic dataset.}
	\label{fig:effect_paras}
\end{figure}

\begin{table}[t]
	\centering
	\setlength{\tabcolsep}{1.175mm}
	\caption{The effect of quantified transferability on medical endoscopic dataset in terms of mIoU, where the last row is the $p$ value of t-test of comparison methods.}
	\scalebox{0.90}{
		\begin{tabular}{|c|ccc|c|}
			\hline
			& ADV \cite{Vu_2019_CVPR} & SIBAN \cite{Luo_2019_ICCV} & DPR \cite{Tsai_2019_ICCV} & ~~Ours~~ \\
			\hline
			without quantified transferability & 63.05 & 61.92 & 63.24 & \textbf{64.17} \\
			with quantified transferability & 63.41 & 62.24 & 63.52 & \textbf{64.43} \\
			t-test ($p$ value) & 0.0014 & 0.0023 & 0.0042 & 0.0031 \\
			\hline				
		\end{tabular}
	}				
	\label{table:exp_QT_medical}
\end{table}

\subsubsection{Ablation Study} This subsection investigates the effectiveness of different components of our proposed model. Fig.~\ref{figure:ablation_exp_medical} reports the performance in term of mIoU achieved by different variants settings on medical dataset. From the presented results in Fig.~\ref{figure:ablation_exp_medical}, we can notice that the performance degrades heavily after removing one or more components. Specifically, the mIoU drops 0.38\%$\sim$4.42\% when pseudo labels selection, semantic representation transfer or quantified transferability is removed. It validates the designs of different components are reasonable. 
Compared with our conference version \cite{Dong_2019_ICCV}, the quantified transferability module and self-supervised pseudo labels generator 
are	indispensable to boost the semantic knowledge transfer across domains. It further justifies that our model can effectively incorporate confident pseudo labels and transferable dependencies to improve segmentation performance.

\begin{table*}[t]	
	\centering
	\setlength{\tabcolsep}{1.74mm}
	\caption{Comparison experiments on SYNTHIA \cite{data:synthia} $\rightarrow$ Cityscapes \cite{data:city} task, where $\mr{mIoU}^{*}$ denotes the $\mr{mIoU}$ for the remaining 13 classes.}
	\scalebox{0.915}{
		\begin{tabular}{|c|cccccccccccccccc|cc|}
			\hline
			Method & road & sidewalk & building & wall & fence & pole &light & sign & veg & sky & person & rider & car & bus & mbike & bike & $\mr{mIoU}$ & $\mr{mIoU}^{*}$ \\
			\hline			
			\hline			
			
			Wild \cite{exp:Wild} & 11.5 & 19.6 & 30.8 & 4.4 & 0.0 & 20.3 & 0.1 & 11.7& 42.3 & 68.7 & 51.2 & 3.8 & 54.0 & 3.2 & 0.2 & 0.6 & 20.2 & - \\
			
			CL \cite{exp:CL} & 65.2 & 26.1 & 74.9 & 0.1 & 0.5 & 10.7& 3.7 & 3.0 &76.1 & 70.6 & 47.1 & 8.2 & 43.2 & 20.7 & 0.7 & 13.1 & 29.0 & - \\ 	
			
			NMD \cite{exp:CCA} & 62.7 & 25.6 & 78.3 & - & - & - & 1.2 & 5.4 & 81.3 & 81.0 & 37.4 & 6.4 & 63.5 &10.1 & 1.2 & 4.6 & - & 35.3 \\
			
			LSD \cite{exp:LSD} &80.1 &29.1 &77.5 &2.8 &0.4 &26.8 &11.1 &18.0 &78.1 &76.7 &48.2 & 15.2&70.5 &17.4 &8.7 &16.7 &36.1 & -\\
			
			LtA \cite{exp:LtA} & 84.3 & \textcolor{blue}{42.7} & 77.5& -& -& -& 4.7 & 7.0& 77.9 & 82.5 & 54.3 & 21.0 & 72.3 & 32.2 & 18.9 & 32.3 & - & 46.7 \\
			
			MCD \cite{Saito_2018_CVPR} & 84.8 & \textcolor{red}{\textbf{43.6}} & 79.0 & 3.9 & 0.2 & 29.1 & 7.2 & 5.5 & \textcolor{blue}{83.8} & \textcolor{blue}{83.1} & 51.0 & 11.7 & \textcolor{blue}{79.9} & 27.2 & 6.2 & 0.0 & 37.3 & - \\

			CGAN \cite{exp:CGAN} & \textcolor{blue}{85.0} & 25.8 & 73.5 & 3.4 & \textcolor{red}{\textbf{3.0}} & 31.5 &19.5 & 21.3 & 67.4 & 69.4 & \textcolor{red}{\textbf{68.5}} & 25.0 & 76.5 &  \textcolor{red}{\textbf{41.6}} & 17.9 & 29.5 & 41.2 & -\\	
			
			DCAN \cite{Wu_2018_ECCV} & 82.8 & 36.4 & 75.7 & 5.1 & 0.1 & 25.8 & 8.0 & 18.7 & 74.7 & 76.9 & 51.1 & 15.9 & 77.7 & 24.8 & 4.1 & 37.3 & 38.4  & -\\
			
			CBST \cite{Zou_2018_ECCV} & 53.6 & 23.7 & 75.0 & 12.5 & 0.3 & \textcolor{red}{\textbf{36.4}} & 23.5 & 26.3 & \textcolor{red}{\textbf{84.8}} & 74.7 & \textcolor{blue}{67.2} & 17.5 & \textcolor{red}{\textbf{84.5}} & 28.4 & 15.2 & \textcolor{red}{\textbf{55.8}} & 42.5 & - \\
			
			CLAN \cite{Luo_2019_CVPR} & 81.3 & 37.0 & \textcolor{blue}{80.1} & - & - & - & 16.1 & 13.7 & 78.2 & 81.5 & 53.4 & 21.2 & 73.0 & 32.9 & 22.6 & 30.7 & - & 47.8 \\
			
			SWD \cite{Lee_2019_CVPR} & 82.4 & 33.2 &\textcolor{red}{\textbf{82.5}} & - & - & - & 22.6 & 19.7 & 83.7 & 78.8 & 44.0 & 17.9 & 75.4 & 30.2 & 14.4 & 39.9 & - & 48.1 \\
			
			ADV \cite{Vu_2019_CVPR} & \textcolor{red}{\textbf{85.6}} & 42.2 & 79.7 & 8.7 & 0.4 & 25.9 & 5.4 & 8.1 & 80.4 & \textcolor{red}{\textbf{84.1}} & 57.9 & 23.8 & 73.3 & \textcolor{blue}{36.4} & 14.2 & 33.0 & 41.2 & - \\
			
			DPR \cite{Tsai_2019_ICCV} & 82.4 & 38.0 & 78.6 & 8.7 & 0.6 & 26.0 & 3.9 & 11.1 & 75.5 & 84.6 & 53.5 & 21.6 & 71.4 & 32.6 & 19.3 & 31.7 & 40.0 & - \\
			
			SIBAN \cite{Luo_2019_ICCV} & 82.5 & 24.0 & 79.4 & - & - & - & 16.5 & 12.7 & 79.2 & 82.8 & 58.3 & 18.0 & 79.3 & 25.3 & 17.6 & 25.9 & - & 46.3 \\
			
			MSL \cite{Chen_2019_ICCV} & 82.9 & 40.7 & 80.3 & 10.2 & 0.8 & 25.8 & 12.8 & 18.2 & 82.5 & 82.2 & 53.1 & 18.0 & 79.0 & 31.4 & 10.4 & 35.6 & 41.4 & - \\
			
			CRST-LR \cite{Zou_2019_ICCV} & 65.6& 30.3& 74.6& 13.8& \textcolor{blue}{1.5} &\textcolor{blue}{35.8}& 23.1& \textcolor{blue}{29.1}& 77.0& 77.5& 60.1 &\textcolor{blue}{28.5}& 82.2& 22.6& 20.1& 41.9 &42.7& - \\
			
			CRST-MR \cite{Zou_2019_ICCV} & 63.4 &27.1& 76.4& 14.2& 1.4& 35.2& 23.6& \textcolor{red}{\textbf{29.4}} &78.5& 77.8& 61.4& \textcolor{red}{\textbf{29.5}}& 82.2& 22.8& 18.9& 42.3& 42.8& - \\
			
			SWE \cite{Dong_2019_ICCV} & 68.4 & 30.1 & 74.2 & 21.5 & 0.4 & 29.2 & 29.3 & 25.1 & 80.3 & 81.5 & 63.1 & 16.4 & 75.6 & 13.5 & 26.1 & \textcolor{blue}{51.9} & \textcolor{blue}{42.9} & -  \\	
			
			\hline
			\hline		
			BL & 22.5 & 15.4 & 74.1 & 9.2 & 0.1 & 24.6 & 6.6 & 11.7 & 75.0 & 82.0 &56.5 & 18.7 & 34.0 & 19.7 & 17.1 & 18.5 & 30.4 & 34.8 \\				
			
			BL+AL & 74.4 & 30.5 & 75.8 & 13.2 & 0.2 & 19.7 & 4.4 &4.9 & 78.2 & 82.7 &44.4 &16.0 & 63.2 & 33.3 & 13.5 &26.2 & 36.3 & 42.1 \\
			
			BL+AL+PL & 79.5 & 39.4 & 77.3 & 11.2 & 0.3 & 22.5 & 5.3 & 11.8 & 79.2 & 81.8 & 58.6 & 20.3 & 71.2 & 31.0 & 25.3 & 33.1 & 40.5 & 47.2 \\
			
			BL+AL+SRT & 79.9 & 38.2 &77.1 & 9.7 & 0.2& 21.1 &6.8 &7.6 &76.1 &81.6 &54.8 & 21.3 &66.2 &30.8 &21.6 &30.6 & 39.0 & 45.6 \\
			
			BL+PL+SRT & 62.8 & 28.5 & 71.3 & 21.3 & 0.6 & 29.2 & \textcolor{blue}{32.1} & 25.3 & 80.7 & 81.1 & 62.3 & 15.7 & 68.9 & 12.6 & 27.6 & 51.7 & 42.0 & 47.8 \\			
			
			BL+AL+PL+QT & 80.3 & 39.6 & 76.8 & 13.7 & 0.5 & 22.3 & 5.8 & 12.1 & 78.7 & 82.2 & 58.4 & 19.6 & 71.9 & 31.3 & 26.5 & 34.1 & 40.9 & 47.5 \\
			
			BL+AL+SRT+QT & 80.7 & 38.5 & 77.4 & 9.2 & 0.4 & 20.7 & 6.5 & 7.3 &78.5 &82.3 &54.1 & 21.8 & 67.5 & 31.7 & 21.2 & 32.4 & 39.4 & 46.2 \\
			
			BL+PL+SRT+QT & 67.3 & 31.5 & 76.9 & 21.4 & 0.5 & 28.1 & 31.9 & 24.2 & 78.4 & 80.2 & 61.4 & 15.9 & 70.9 & 13.2 & 24.1 & 49.2 & 42.2 & 48.1 \\
			
			Ours-woSP & 70.8 & 32.9 & 76.1 & \textcolor{blue}{21.6} & 0.6 & 28.9 & 30.5 & 23.6 & 77.2 & 78.8 & 62.3 & 16.4 & 72.5 & 12.1 & \textcolor{red}{\textbf{28.6}} & 49.4 & 42.6 & \textcolor{blue}{48.5} \\
			
			Ours & 70.3 & 32.1 & 75.7 & \textcolor{red}{\textbf{22.9}} & 0.8 & 29.6 & \textcolor{red}{\textbf{32.4}} & 24.3 & 79.7 & 78.1 & 62.5 & 17.0 & 75.0 & 13.2 & \textcolor{blue}{28.3} & 51.7 & \textcolor{red}{\textbf{43.4}} & \textcolor{red}{\textbf{49.3}} \\		
			\hline 		
		\end{tabular}
	}		
	\label{table:exp_synthia_cityscapes}
\end{table*}

\subsubsection{Qualitative Analysis About Pseudo Labels} To verify the importance of our self-supervised pseudo label generator with class balance and spatial continuity, we conduct comparison experiments with other ablation designs, \emph{i.e.}, optimizing without pseudo labels (Our-woPL), optimizing without super-pixel spatial prior (Ours-woSP), optimizing without class balance (Ours-woCB) and optimizing without self-supervision (Ours-woSS). As shown in Fig.~\ref{fig:ablation_pseudo_label}, compared with Ours-woPL, our proposed pseudo label generator achieves 2.36\% improvement in terms of mIoU by incorporating class balance and spatial prior in a self-supervised manner. 
Besides, it performs better than our conference version SWE \cite{Dong_2019_ICCV}, since the quantified transferability module and self-supervised training manner can explore transferable representations and prevent deviation of false pseudo labels in the retraining phase.
Furthermore, Fig.~\ref{fig:pseudo_label_propagation_medical} intuitively demonstrates the propagation process of pseudo labels, where our model could iteratively produce more confident pseudo labels for fine-tuning process.

\subsubsection{Effect of Hyper-Parameters} We intend to study the parameter effects with respect to $\{\mu, \eta\}$ and $\{\alpha, \gamma\}$ in this subsection. The optimal values of hyper-parameters are determined by empirically conducting extensive parameter experiments, as presented in Fig.~\ref{fig:effect_paras}. We set $\eta=0.3, \mu=10, \alpha=1$ and $\gamma=0.7$ for best performance in all experiments. Moreover, our proposed model achieves stable transfer performance on the medical endoscopic dataset even though there is a wide selection range for hyper-parameters. The results of Fig.~\ref{fig:effect_paras} (b) justify that incorporating previously-learned features and sparsity property for different feature centroids is essential to improve the performance of our model.

\subsubsection{Effect of Quantified Transferability} This subsection validates the effectiveness of quantified transferability when combined with other state-of-the-art models, such as ADV \cite{Vu_2019_CVPR}, SIBAN \cite{Luo_2019_ICCV} and DPR \cite{Tsai_2019_ICCV}, as presented in Table~\ref{table:exp_QT_medical}. From the presented results, we observe that: 1) The performance of comparison methods equipped with our proposed quantified transferability improves 0.28\%$\sim$0.36\% mIoU on medical endoscopic datasets. It illustrates that the quantified transferability module could explore transferable knowledge while preventing negative transfer of untransferable knowledge. 2) Our model outperforms these competing methods even with quantified transferability about 0.91\% $\sim$ 2.19\% mIoU, which further verifies the superiority of our model.

To further validate the effectiveness of quantified transferability, we conduct the t-test experiments on all datasets to evaluate whether the improvement with quantified transferability mechanism is significant, as shown in Table~\ref{table:exp_QT_medical}. Specifically, the lower value of $p$ in Table~\ref{table:exp_QT_medical} is, the more confident the significant improvement between the comparison methods with quantified transferability will be. We run each comparison method for 5 times, and then conduct the t-test evaluation. Obviously, the presented results (\emph{i.e.}, $p\textless$0.05) effectively demonstrate the significance of the improvements of comparison methods with quantified transferability.

\begin{figure}[t]
	\centering
	\includegraphics[trim = 0mm 0mm 0mm 0mm, clip, width = 240pt, height =170pt]{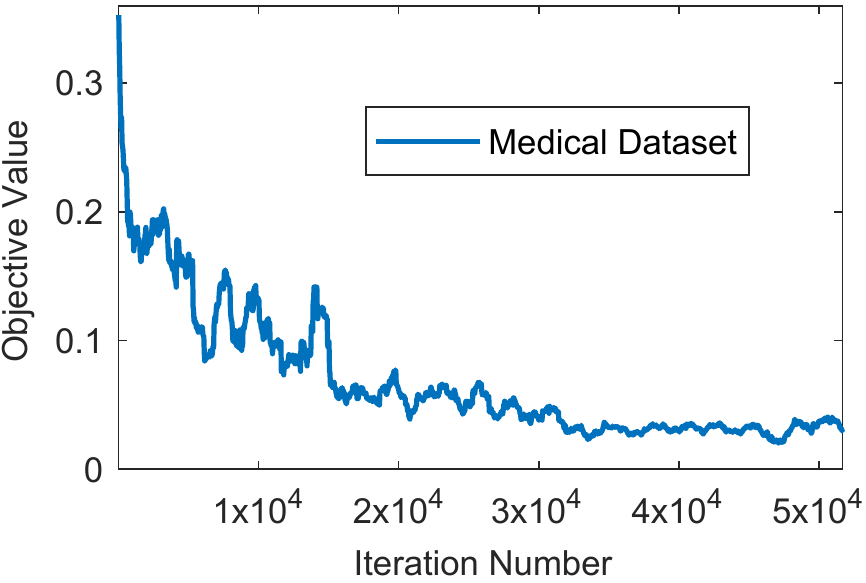}
	\caption{Convergence analysis on medical endoscopic dataset.}
	\label{fig:converge_medical}
\end{figure}

\begin{figure}[t]
	\centering
	\includegraphics[trim = 0mm 0mm 0mm 0mm, clip, width = 240pt, height =170pt]{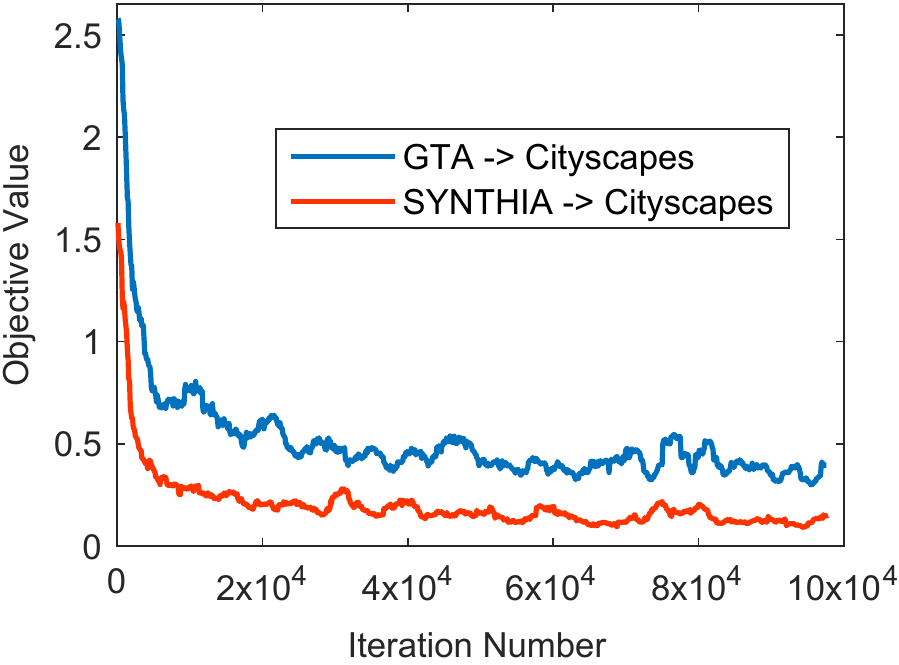}
	\caption{Convergence analysis of our model on GTA $\rightarrow$ Cityscapes and SYNTHIA $\rightarrow$ Cityscapes adaptation tasks. }
	\label{fig:converge_pulic}
\end{figure}

\begin{table*}[t]
	\centering
	\setlength{\tabcolsep}{1.185mm}
	\caption{Comparison experiments on GTA \cite{data:GTA} $\rightarrow$ Cityscapes \cite{data:city} transfer task.}
	\scalebox{0.94}{
		\begin{tabular}{|c|ccccccccccccccccccc|c|}
			\hline
			Method & road & sidewalk & building & wall & fence & pole &light &sign & veg & terrain & sky  & person & rider & car & truck & bus & train & mbike & bike & mIoU \\	
			\hline
			\hline
			
			Wild \cite{exp:Wild} & 70.4 & 32.4 & 62.1 & 14.9 & 5.4 & 10.9 & 14.2 & 2.7 & 79.2& 21.3 & 64.6 & 44.1 & 4.2 & 70.4 & 8.0 & 7.3& 0.0 & 3.5 & 0.0 & 27.1 \\
			
			CL \cite{exp:CL} & 74.9 & 22.0 & 71.7 & 6.0 & 11.9 & 8.4 & 16.3 & 11.1 & 75.7 & 11.3 & 66.5 & 38.0 & 9.3 & 55.2 & 18.8 & 18.9 & 0.0 & 16.8 & 14.6 & 28.9\\
			
			CyCADA \cite{domain:class-preserve-2} &79.1 &33.1 &77.9 &23.4 &17.3 &32.1 &33.3 &31.8 &81.5 &26.7 & 69.0 &\textcolor{blue}{62.8} &14.7 &74.5 &20.9 &25.6 & 6.9&18.8 &20.4 & 39.5 \\
			
			LSD \cite{exp:LSD} &88.0 &30.5 &78.6 &25.2 &23.5& 16.7 &23.5 &11.6 &78.7 &27.2 &71.9 &51.3 &19.5 &80.4 &19.8 &18.3 &0.9 &20.8 &18.4 &37.1 \\
			
			LtA \cite{exp:LtA} & 86.5 & 36.0 & 79.9 & 23.4 & 23.3 &23.9 &35.2 &14.8 & 83.4& 33.3 & 75.6 & 58.5 & 27.6 & 73.7 & 32.5 & 35.4 & 3.9 & 30.1 & 28.1 & 42.4 \\
			
			MCD \cite{Saito_2018_CVPR} & 90.3 & 31.0 & 78.5 & 19.7 & 17.3 & 28.6 & 30.9 & 16.1 & \textcolor{blue}{83.7} & 30.0 & 69.1 & 58.5 & 19.6 & 81.5 & 23.8 & 30.0 & 5.7 &25.7 & 14.3 & 39.7  \\
			
			CGAN \cite{exp:CGAN} & 89.2 & 49.0 & 70.7&13.5 & 10.9& 38.5 & 29.4 & 33.7& 77.9 & \textcolor{red}{\textbf{37.6}} & 65.8 & \textcolor{red}{\textbf{75.1}} & \textcolor{red}{\textbf{32.4}}& 77.8 & \textcolor{red}{\textbf{39.2}}&45.2 & 0.0& 25.2 & \textcolor{blue}{35.4} & 44.5 \\
			
			DLOW \cite{Gong_2019_CVPR} & 87.1 & 33.5 & 80.5 & 24.5 & 13.2 & 29.8 & 29.5 & 26.6 & 82.6 & 26.7 & \textcolor{red}{\textbf{81.8}} & 55.9 & 25.3 & 78.0 & 33.5 & 38.7 & 0.0 & 22.9 & 34.5 & 42.3  \\
			
			CBST \cite{Zou_2018_ECCV} & 88.0 & \textcolor{red}{\textbf{56.2}} & 77.0 & 27.4 & 22.4 & 40.7 & 47.3 & 40.9 & 82.4 & 21.6 & 60.3 & 50.2 & 20.4 & 83.8 & 35.0 & \textcolor{red}{\textbf{51.0}} & 15.2 & 20.6 & 37.0 & 46.2  \\
			
			CLAN \cite{Luo_2019_CVPR} & 87.0 & 27.1 & 79.6 & 27.3 & 23.3 & 28.3 & 35.5 & 24.2 & 83.6 & 27.4 & 74.2 & 58.6 & 28.0 & 76.2 & 33.1 & 36.7 & 6.7 & 31.9 & 31.4 & 43.2  \\
			
			SWD \cite{Lee_2019_CVPR} & 92.0 & 46.4 & \textcolor{red}{\textbf{82.4}} & 24.8 & 24.0 & 35.1 & 33.4 & 34.2 & 83.6 & 30.4 & \textcolor{blue}{80.9} & 56.9 & 21.9 & 82.0 & 24.4 & 28.7 & 6.1 & 25.0 & 33.6 & 44.5 \\
			
			ADV \cite{Vu_2019_CVPR} & 89.4 & 33.1 & 81.0 & 26.6 & 26.8 & 27.2 & 33.5 & 24.7 & \textcolor{red}{\textbf{83.9}} & \textcolor{blue}{36.7} & 78.8 & 58.7 & 30.5 & 84.8 & \textcolor{blue}{38.5} & 44.5 & 1.7 & 31.6 & 32.5 & 45.5 \\
			
			DPR \cite{Tsai_2019_ICCV} & 92.3 & 51.9 & \textcolor{blue}{82.1} & 29.2 & 25.1 & 24.5 & 33.8 & 33.0 & 82.4 & 32.8 & 82.2 & 58.6 & 27.2 & 84.3 & 33.4 & 46.3 & 2.2 & 29.5 & 32.3 & 46.5  \\
			
			SIBAN \cite{Luo_2019_ICCV}& 88.5 & 35.4 & 79.5 & 26.3 & 24.3 & 28.5 & 32.5 & 18.3 & 81.2 & 40.0 & 76.5 & 58.1 & 25.8 & 82.6 & 30.3 & 34.4 & 3.4 & 21.6 & 21.5 & 42.6 \\
			
			MSL \cite{Chen_2019_ICCV} & 89.4 & 43.0 & \textcolor{blue}{82.1} & \textcolor{blue}{30.5} & 21.3 & 30.3 & 34.7 & 24.0 & 85.3 & 39.4 & 78.2 & 63.0 & 22.9 & 84.6 & 36.4 & 43.0 & 5.5 & 34.7 & 33.5 & 46.4  \\
			
			CRST-LR \cite{Zou_2019_ICCV} & 80.3& 40.8& 65.8& 24.6& \textcolor{blue}{30.5}& \textcolor{blue}{43.1}& \textcolor{blue}{49.5}& 40.3& 82.1& 26.0& 54.6& 59.4& \textcolor{blue}{32.1}& 68.0& 31.9& 30.0& 21.9& \textcolor{red}{\textbf{44.8}}& \textcolor{red}{\textbf{46.7}}& 45.9  \\
			
			CRST-MR \cite{Zou_2019_ICCV} & 84.4& \textcolor{blue}{52.7}& 74.7& \textcolor{red}{\textbf{38.0}}& \textcolor{red}{\textbf{32.2}}& \textcolor{red}{\textbf{43.7}}& \textcolor{red}{\textbf{53.7}}& 38.6& 73.9& 24.4& 64.4& 45.6& 24.6& 63.2& 3.22& 31.9& \textcolor{red}{\textbf{45.9}}& \textcolor{blue}{44.2}& 34.8& 46.0  \\
			
			SWE \cite{Dong_2019_ICCV} &\textcolor{blue}{92.7} &48.0 &78.8 & 25.7 & 27.2 & 36.0 & 42.2 & 45.3& 80.6 &14.6 &66.0 & 62.1& 30.4 &86.2 & 28.0 &45.6 &35.9 &16.8 & 34.7 & \textcolor{blue}{47.2} \\
			
			\hline	
			\hline 		
			BL & 80.2 & 6.4 & 74.8 & 8.8 & 17.2 & 17.5 & 30.5 & 17.7 & 75.0 & 14.1 & 57.9 &56.2 & 27.3 &64.1 & 29.7 & 24.1 & 4.7 & 27.6 & 33.4 & 35.1  \\  				
			
			BL+AL & 86.3 & 32.2 & 79.8 &22.0 &22.2 &27.1 &33.5 &20.1 &80.3 & 21.5 &75.5 &59.0 &25.4 &73.1 & 28.0 &32.2 & 5.4 & 27.3& 31.5& 41.2 \\
			
			BL+AL+PL & 91.3 & 48.7 & 77.6 & 23.9 & 28.7 & 30.4 & 41.8 & 43.0 & 77.8 & 15.4 & 73.5 & 55.6 & 19.8 & 86.3 & 22.5 & 43.2 & 34.5 & 14.6 & 30.8 & 45.2  \\
			
			BL+AL+SRT & 92.4 & 49.8 &73.6 &25.3 & 28.3 &24.5 &40.9 &45.0 &79.2 &14.2 & 70.4&50.1 & 18.6 & \textcolor{blue}{86.6} & 22.3 & 45.4 & 30.3 & 11.9 & 32.8 & 44.3 \\
			
			BL+PL+SRT & 92.4 & 48.3 & 75.6 & 27.5 & 27.3 &30.5 & 42.7 &\textcolor{blue}{45.9} & 81.7 & 15.8 & 73.0 & 56.5 & 25.8 &86.1 & 23.6 & 41.7 & \textcolor{blue}{43.8} &13.3 & 30.9 & 46.4  \\
			
			BL+AL+PL+QT & 92.1 & 49.4 & 78.3 & 22.6 & 26.1 & 32.7 & 42.3 & 41.8 & 79.5 & 16.1 & 72.7 & 57.8 & 17.4 & 85.6 & 24.8 & 42.5 & 36.2 & 15.1 & 31.4 & 45.5  \\
			
			BL+AL+SRT+QT & 92.2 & 50.4 & 75.1 & 24.6 & 27.5 & 27.2 &39.6 & 44.1 & 80.3 & 15.7 & 72.8 & 55.2 & 16.5 & 85.3 & 21.4 & 43.9 & 33.6 & 12.3 & 31.1 & 44.7  \\
			
			BL+PL+SRT+QT & 92.5 & 48.3 & 73.8 & 27.8 & 26.1 & 28.9 &44.3 & 45.2 & 80.4 & 17.6 & 73.5 & 62.7 & 24.1 &84.7 & 27.0 & 42.6 & 41.5 & 16.4 & 31.3 & 46.8  \\
			
			Ours-woSP & 92.5 & 47.9 & 77.1 & 24.9 & 27.3 & 35.4 & 41.5 & 43.8 & 78.2 & 18.7 & 72.4 & 64.3 & 31.5 & 83.8 & 27.9 & 43.4 & 32.7 & 18.5 & 33.6 & 47.1 \\
			
			Ours &\textcolor{red}{\textbf{92.8}} & 48.6 & 79.2 & 26.2 & 26.4 & 36.8 & 40.6 & \textcolor{red}{\textbf{47.1}} & 80.5 & 19.3 &69.0 & 63.1& 28.9 &\textcolor{red}{\textbf{86.7}} & 29.2  &\textcolor{blue}{45.8} &34.3 & 19.4  &32.7 & \textcolor{red}{\textbf{47.7}} \\		
			\hline 		
		\end{tabular}
	}			
	\label{table:exp_gta_cityscapes}
\end{table*}

\begin{table}[t]
	\centering
	\setlength{\tabcolsep}{1.175mm}
	\caption{The effect of quantified transferability on SYNTHIA $\rightarrow$ Cityscapes (mIoU), where the last row is the $p$ value of t-test.}
	\scalebox{0.90}{
		\begin{tabular}{|c|ccc|c|}
			\hline
			& ADV \cite{Vu_2019_CVPR} & SIBAN \cite{Luo_2019_ICCV} & DPR \cite{Tsai_2019_ICCV} & ~~Ours~~ \\
			\hline
			without quantified transferability & 41.2 & 37.4 & 40.0 &  \textbf{43.1} \\
			with quantified transferability & 41.6 & 37.7 & 40.5 & \textbf{43.4} \\
			t-test ($p$ value) & 0.0051 & 0.0063 & 0.0042 & 0.0048  \\
			\hline				
		\end{tabular}
	}				
	\label{table:exp_QT_SYN}
\end{table}

\begin{table}[t]
	\centering
	\setlength{\tabcolsep}{1.175mm}
	\caption{The effect of quantified transferability on GTA $\rightarrow$ Cityscapes in terms of mIoU, where the last row is the $p$ value of t-test.}
	\scalebox{0.90}{
		\begin{tabular}{|c|ccc|c|}
			\hline
			& ADV \cite{Vu_2019_CVPR} & SIBAN \cite{Luo_2019_ICCV} & DPR \cite{Tsai_2019_ICCV} & ~~Ours~~ \\
			\hline
			without quantified transferability & 45.5 & 42.6 & 46.5 &  \textbf{47.5} \\
			with quantified transferability & 45.9 & 42.8 & 46.8 & \textbf{47.7} \\
			t-test ($p$ value) & 0.0047 & 0.0084 & 0.0061 & 0.0096 \\
			\hline				
		\end{tabular}
	}				
	\label{table:exp_QT_GTA}
\end{table}

\subsubsection{Convergence Analysis} This subsection introduces the convergence analysis about our proposed model. As shown in Fig.~\ref{fig:converge_medical}, our model could achieve efficient convergence (\emph{i.e.}, iteration number is about $3.5 \times 10^{4}$) along the iterative training process. 
The experimental convergence curve effectively supports the theoretical analysis in Section \ref{sec:theoretical_insights}, \emph{i.e.}, our model could narrow the domain discrepancy $d_{\mathcal{H}\triangle\mathcal{H}(P_s, P_t)}$ and minimize the upper bound of target expected error $\epsilon_T(h)$.

\subsection{Experiments on Benchmark Datasets} 	
In this subsection, to further verify the generalization and robustness of our model, we conduct extensive comparison experiments on several representative public datasets (\emph{i.e.}, SYNTHIA \cite{data:synthia} $\rightarrow$ Cityscapes \cite{data:city} and GTA \cite{data:GTA} $\rightarrow$ Cityscapes \cite{data:city}). To be specific, we remove the image classifier to neglect its assistance for pixel predictions, and utilize the same experiment settings with other comparison approaches \cite{exp:LSD, exp:LtA, exp:CGAN, Vu_2019_CVPR, Dong_2019_ICCV, Lee_2019_CVPR}.

The convergence analysis of our model on benchmark datasets are depicted in Fig.~\ref{fig:converge_pulic}, which validates that our model achieves tight upper bound for target expected error $\epsilon_T(h)$ along the iterative training process. The experimental results in Fig.~\ref{fig:converge_pulic} well match the theoretical analysis in Section \ref{sec:theoretical_insights}.

\subsubsection{Transfer from SYNTHIA to Cityscapes} Learning transferable dependencies from SYNTHIA \cite{data:synthia} to Cityscapes \cite{data:city} is our focus in this experiment. In the training stage, we regard finely-labeled 9400 samples from SYNTHIA and 2993 samples from Cityscapes without pixel labels as $X^s$ and $X^t$, respectively. For the evaluation, validation subset with 500 images from Cityscapes dataset is utilized, where this subset is disjoint with the training data. As shown in  Table~\ref{table:exp_synthia_cityscapes}, 16 shared classes for SYNTHIA and Cityscapes datasets are considered. From  Table~\ref{table:exp_synthia_cityscapes}, we can notice that: 
1) Our model improves the transfer performance of $0.4\%\sim6.0\%$ than competing models \cite{Luo_2019_CVPR, Lee_2019_CVPR, Vu_2019_CVPR, Zou_2018_ECCV, Saito_2018_CVPR} and our conference version \cite{Dong_2019_ICCV} for 16 categories in terms of mIoU, which validates our proposed model can efficiently explore transferable dependencies across source and target datasets while neglecting the irrelevant knowledge; 
2) Ablation performance further justifies each designed component play an important role in narrowing the semantic gap; 
3) When compared with \cite{Dong_2019_ICCV}, our model has comparable performance for hard-to-transfer categories with extremely different appearances across domains (\emph{e.g.}, wall, pole, light, motorbike and rider), since pseudo label generator generates confident pixel labels for target samples by considering both class balance and super-pixel prior in a self-supervised manner.
Furthermore, as presented in Table~\ref{table:exp_QT_SYN}, the quantified transferability module is combined with some state-of-the-art models, \emph{i.e.}, ADV \cite{Vu_2019_CVPR}, SIBAN \cite{Luo_2019_ICCV} and DPR \cite{Tsai_2019_ICCV} to further demonstrate its scale application ability for other models. From the t-test evaluation results in Table~\ref{table:exp_QT_SYN}, we could observe that the competing models with quantified transferability improve the transfer performance significantly.

\subsubsection{Transfer from GTA to Cityscapes} 
For model training, we employ GTA \cite{data:GTA} with finely-annotated 24996 samples as source dataset $X^s$, and utilize the training subset of Cityscapes \cite{data:city} with unlabeled 2993 samples as target dataset $X^t$. Besides, we use the validation subset of Cityscapes \cite{data:city} including 500 samples to evaluate the performance. Table~\ref{table:exp_gta_cityscapes} reports the results for 19 common classes. From Table~\ref{table:exp_gta_cityscapes}, we have the following observations: 
1) Our model pays equal attention for both easy-to-transfer categories and initially hard-to-transfer classes, while other competing methods \cite{exp:Wild, exp:LSD, exp:LtA, exp:CGAN} give priority to those well-to-transfer classes with large amount of pixels (\emph{e.g.}, vegetation, building, road, car and sky).
2) Ablation experiments validate that each developed module reasonable designs are efficient to improve semantic transfer performance.
3) Compared with our conference paper \cite{Dong_2019_ICCV}, our model with the Wasserstein quantified transferability focuses on those transferable semantic knowledge and simultaneously ignores the irrelevant regions (\emph{e.g.}, background and some invalid classes).
4) The self-supervised pseudo label generator produces highly-confident pseudo labels, which assists the feature distributions among source and target datasets in both global and local views aligned better.   
Moreover, as shown in Table~\ref{table:exp_QT_GTA}, we apply the quantified transferability module into other competing models, \emph{i.e.}, ADV \cite{Vu_2019_CVPR}, SIBAN \cite{Luo_2019_ICCV} and DPR \cite{Tsai_2019_ICCV} to further validate its application advantage and the significant improvement for knowledge transfer performance.

\section{Conclusion} \label{sec:conclusion}
In this paper, we present a new weakly-supervised lesions transfer framework, which highlights important transferable dependencies while neglecting those irrelevant untransferable representations. To be specific, a Wasserstein adversarial learning based quantified transferability framework is developed to selectively capture transferable semantic representations across different domains. Moreover, we propose a self-supervised pseudo pixel label generator under curriculum learning strategy to pick more hard-to-transfer samples from target domain for the model retraining process. It incorporates class balance and super-pixel prior to prevent the enormous deviation of false pseudo labels and the dominance of well-to-transfer classes in a self-supervised manner. With the assistance of dynamically-updated semantic consistence loss, category-wise feature centroids are aligned to narrow the conditional distributions across domains. Extensive comparison experiments on several medical and non-medical domain adaptation datasets justify the robustness and effectiveness of our model. In the future, we will explore unsupervised semantic transfer across gastroscope and enteroscopy datasets for lesions segmentation and apply it into clinical diagnosis applications.

\ifCLASSOPTIONcaptionsoff
  \newpage
\fi

\bibliographystyle{IEEEtran}
\bibliography{Trans_LesionsTransfer}

\begin{IEEEbiography}[{\includegraphics[width=1in,height=1.25in,clip,keepaspectratio]{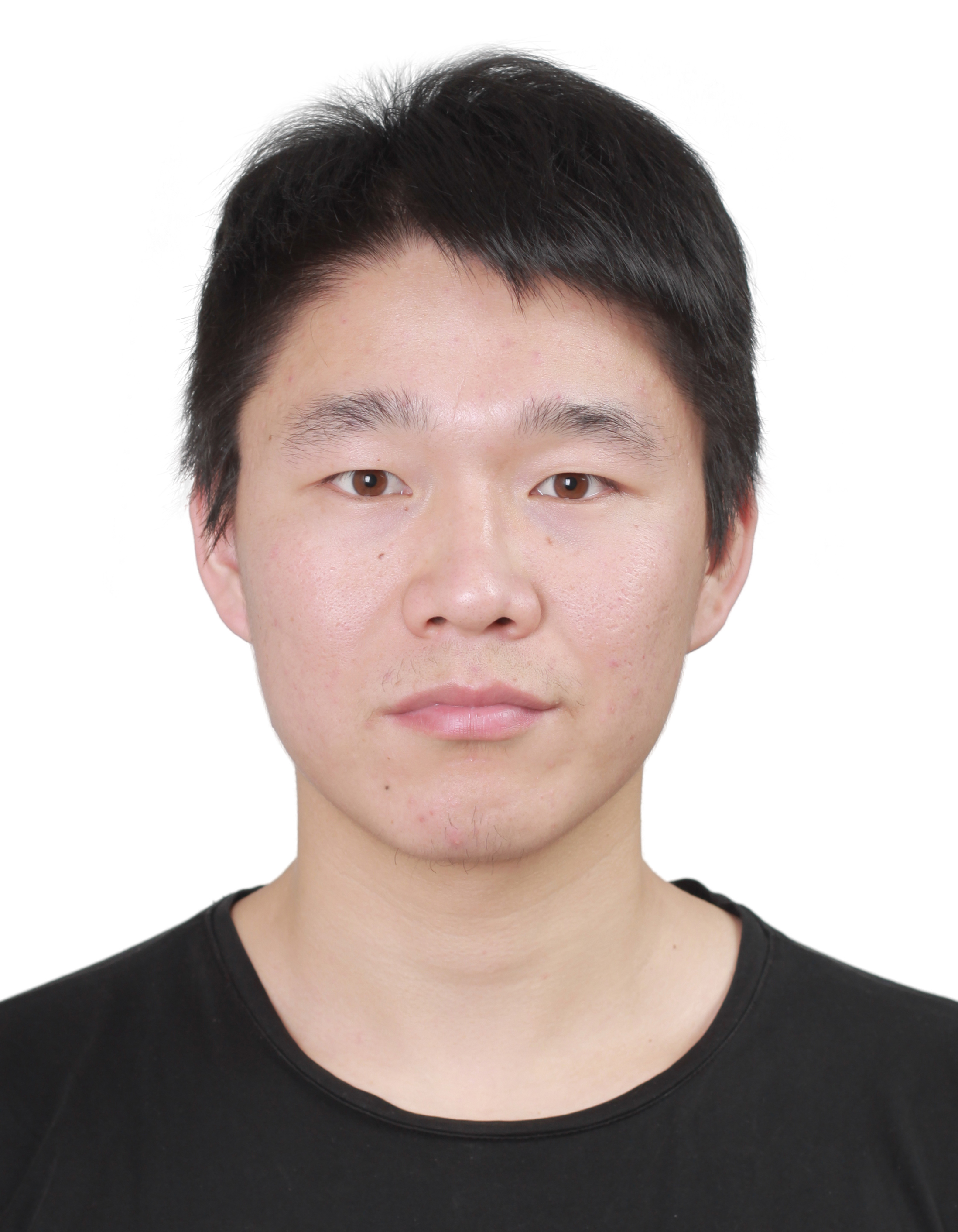}}]{Jiahua Dong} is currently a Ph. D candidate in State Key Laboratory of Robotics, Shenyang Institute of Automation, University of Chinese Academy of Sciences. He received the B.S. degree from Jilin University in 2017. His current research interests include computer vision, machine learning, transfer learning, domain adaptation and medical image processing.
\end{IEEEbiography}

\begin{IEEEbiography}[{\includegraphics[width=1in,height=1.25in,clip,keepaspectratio]{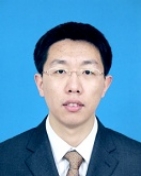}}]{Yang Cong} (S’09-M’11-SM’15) is a full professor of Chinese Academy of Sciences. He received the he B.Sc. de. degree from Northeast University in 2004, and the Ph.D. degree from State Key Laboratory of Robotics, Chinese Academy of Sciences in 2009. He was a Research Fellow of National University of Singapore (NUS) and Nanyang Technological University (NTU) from 2009 to 2011, respectively; and a visiting scholar of University of Rochester. He has served on the editorial board of the Journal of Multimedia. His current research interests include image processing, compute vision, machine learning, multimedia, medical imaging, data mining and robot navigation. He has authored over 70 technical papers. He is also a senior member of IEEE.
\end{IEEEbiography}

\begin{IEEEbiography}[{\includegraphics[width=1in,height=1.25in,clip,keepaspectratio]{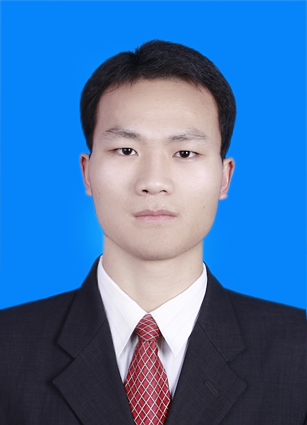}}]{Gan Sun} (S'19) is an Assistant Professor in State Key Laboratory of Robotics, Shenyang Institute of Automation, Chinese Academy of Sciences. He received the B.S. degree from Shandong Agricultural University in 2013, the Ph.D. degree from State Key Laboratory of Robotics, Chinese Academy of Sciences in 2020, and has been visiting Northeastern University from April 2018 to May 2019, Massachusetts Institute of Technology from June 2019 to November 2019. His current research interests include lifelong machine learning, multi-task learning, medical data analysis, deep learning and 3D computer vision.
\end{IEEEbiography}

\begin{IEEEbiography}[{\includegraphics[width=1in,height=1.25in, clip, keepaspectratio]{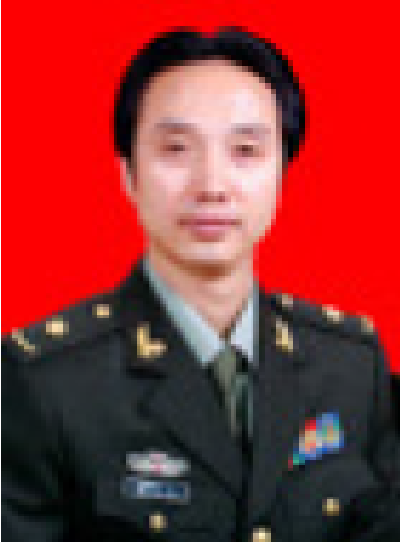}}]{Yunsheng Yang} is currently a professor of Department of Gastroenterology and Hepatology, director of his department for 12 years and Chairman for the Institute of Digestive Diseases in the Chinese PLA General Hospital (301 Hospital) in Beijing. He serves as the supervisor for Ph.D student at Nankai University in Tianjing, Tsinghua University and Chinese PLA Medical Academy in Beijing. Professor Yang graduated from Medical College in 1983, received his master’s degree of medical science from Henan Medical University in 1992 and Ph.D degree from the First Military Medical University in 1995. Professor Yang is the Founder of Sino-NOTES Club and also serves of various of editorial boards and editor-in-chief for Chinese Version of Gut and Chinese Version of Gastroenterology, associated editor-in-chief JDD and Chin J Digestion, editorial member of Gut, and APT, etc. So far, his more than 200 articles have been published, including 150 of original articles in various of international journals, like gastrointestinal endoscopy GIE, hepatology and so on.
\end{IEEEbiography}

\begin{IEEEbiography}[{\includegraphics[width=1in,height=1.25in,clip,keepaspectratio]{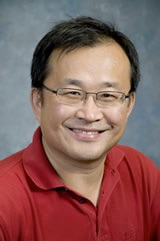}}]{Xiaowei Xu} is a professor of Information Science at University of Arkansas at Little Rock (UALR), received a  a B.Sc. de. degree in Mathematics from Nankai University in 1983 and a Ph.D. degree in Computer Science from University of Munich in 1998. He holds an adjunct professor position in the Department of Mathematics and Statistics at University of Arkansas at Fayetteville. Before his appointment in UALR, he was a senior research scientist in Siemens. He was a visiting professor in Microsoft Research Asia and Chinese University of Hong Kong. His research spans data mining, machine learning, bioinformatics, data management and high performance computing. He has published over 70 papers in peer reviewed journals and conference proceedings. His groundbreaking work on density-based clustering algorithm DBSCAN has been widely used in many textbooks; and received over 10203 citations based on Google scholar. Dr. Xu is a recipient of 2014 ACM KDD Test of Time Award that “recognizes outstanding papers from past KDD Conferences beyond the last decade that have had an important impact on the data mining research community.”
\end{IEEEbiography}

\begin{IEEEbiography}[{\includegraphics[width=1in,height=1.25in, clip, keepaspectratio]{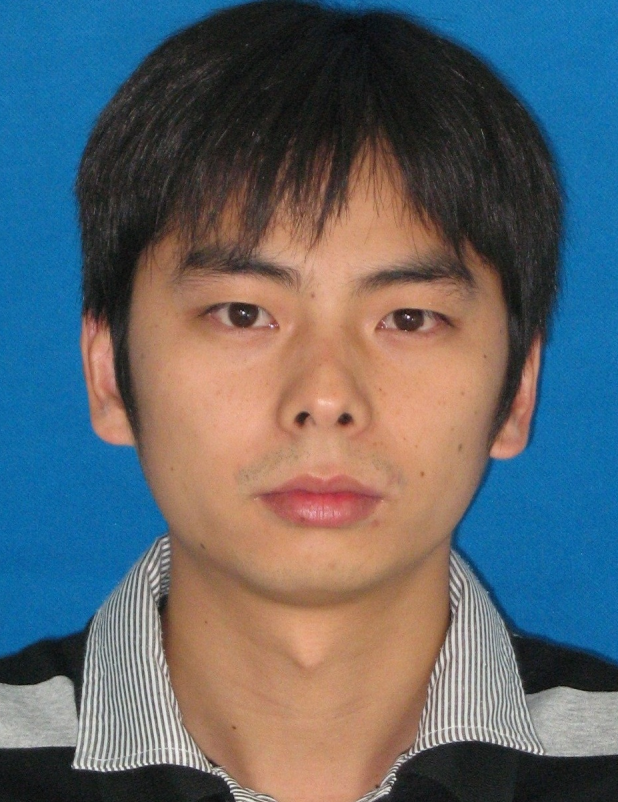}}]{Zhengming Ding} (S’14-M’18) received the B.Eng. degree in information security and the M.Eng. degree in computer software and theory from University of Electronic Science and Technology of China (UESTC), China, in 2010 and 2013, respectively. He received the Ph.D. degree from the Department of Electrical and Computer Engineering, Northeastern University, USA in 2018. He is a faculty member affiliated with Department of Computer, Informationand Technology, Indiana University-Purdue University Indianapolis since 2018. His research interests include transfer learning, multi-view learning and deep learning. He received the National Institute of Justice Fellowship during 2016-2018. He was the recipients of the best paper award (SPIE 2016) and best paper candidate (ACMMM 2017). He is currently an Associate Editor of the Journal of Electronic Imaging (JEI) and IET Image Processing. He is a member of IEEE,  ACM and AAAI.
\end{IEEEbiography}

\end{document}